\documentclass[letterpaper]{article} 
\usepackage{aaai2026}  
\usepackage{times}  
\usepackage{helvet}  
\usepackage{courier}  
\usepackage[hyphens]{url}  
\usepackage{graphicx} 
\usepackage{natbib}  
\usepackage{caption} 
\usepackage{algorithm}
\usepackage{algorithmic}
\usepackage{amsmath, amssymb}
\usepackage{booktabs}
\usepackage{listings}
\usepackage{xcolor}
\usepackage{multirow} 
\usepackage{colortbl}
\definecolor{lightgray}{gray}{0.9}
\usepackage{subcaption}
\usepackage{newfloat}
\usepackage{listings}
\DeclareCaptionStyle{ruled}{labelfont=normalfont,labelsep=colon,strut=off} 
\floatstyle{ruled}
\newfloat{listing}{tb}{lst}{}
\floatname{listing}{Listing}
\title{Format as a Prior:\\ Quantifying and Analyzing Bias in LLMs for Heterogeneous Data}

\author {
    Jiacheng Liu\textsuperscript{\rm 1, \thanks{Jiacheng Liu and Mayi Xu contribute equally to this work.}},
    Mayi Xu\textsuperscript{\rm 1, $^*$},
    Qiankun Pi\textsuperscript{\rm 1}, 
    Wenli Li\textsuperscript{\rm 1}, 
    Ming Zhong\textsuperscript{\rm 1}, 
    Yuanyuan Zhu\textsuperscript{\rm 1},\\
    Mengchi Liu\textsuperscript{\rm 1}, 
    Tieyun Qian\textsuperscript{\rm 1, \thanks{Corresponding author}}
}
\affiliations {
    \textsuperscript{\rm 1}School of Computer Science, Wuhan University, China\\
    \{liu-jia-cheng, xumayi, qty\}@whu.edu.cn
}

\begin{document}

\maketitle

\begin{abstract}

Large Language Models (LLMs) are increasingly employed in applications that require processing information from heterogeneous formats, including texts, tables, infoboxes, and knowledge graphs. However, systematic biases toward particular formats may undermine LLMs' ability to integrate heterogeneous data impartially, potentially resulting in reasoning errors and increased risks in downstream tasks. Yet it remains unclear \emph{whether such biases are systematic}, \emph{which data-level factors drive them}, and \emph{what internal mechanisms underlie their emergence}.
 
In this paper, we present the first comprehensive study of format bias in LLMs through a three-stage empirical analysis. The first stage explores the presence and direction of bias across a diverse range of LLMs. The second stage examines how key data-level factors influence these biases. The third stage analyzes how format bias emerges within LLMs' attention patterns and evaluates a lightweight intervention to test its effectiveness. Our results show that format bias is consistent across model families, driven by information richness, structure quality, and representation type, and is closely associated with attention imbalance within the LLMs. Based on these investigations, we identify three future research directions to reduce format bias: enhancing data pre-processing through format repair and normalization, introducing inference-time interventions such as attention re-weighting, and developing format-balanced training corpora. These directions will support the design of more robust and fair heterogeneous data processing systems.
\end{abstract}

\begin{links}
    \link{Code}{https://github.com/NLPGM/Format-as-a-prior}
    \link{Appendix}{https://github.com/NLPGM/Format-as-a-prior/appendix.pdf}
\end{links}

\section{Introduction}
Large Language Models (LLMs) have demonstrated impressive capabilities across a wide range of natural language tasks~\cite{brown2020language}. However, their practical deployment remains constrained by key limitations, including factual inaccuracies (commonly referred to as “hallucinations”)~\cite{ji2023survey} and incomplete or outdated knowledge~\cite{petroni2019language}. One promising direction to address these issues is to incorporate external knowledge sources into the reasoning process—allowing models to ground their outputs in more accurate, up-to-date, and contextually relevant information~\cite{lewis2020retrieval,gao2023retrieval,huang2022towards}.

In practice, external knowledge exists in diverse formats, ranging from unstructured texts to semi-structured infoboxes, as well as structured tables and Knowledge Graphs (KGs). 
These different sources of knowledge often complement one another, and the ability to effectively harness them together is crucial for real-world, knowledge-intensive applications.
\begin{figure}[tbp]
    \centering
    \includegraphics[width=\columnwidth]{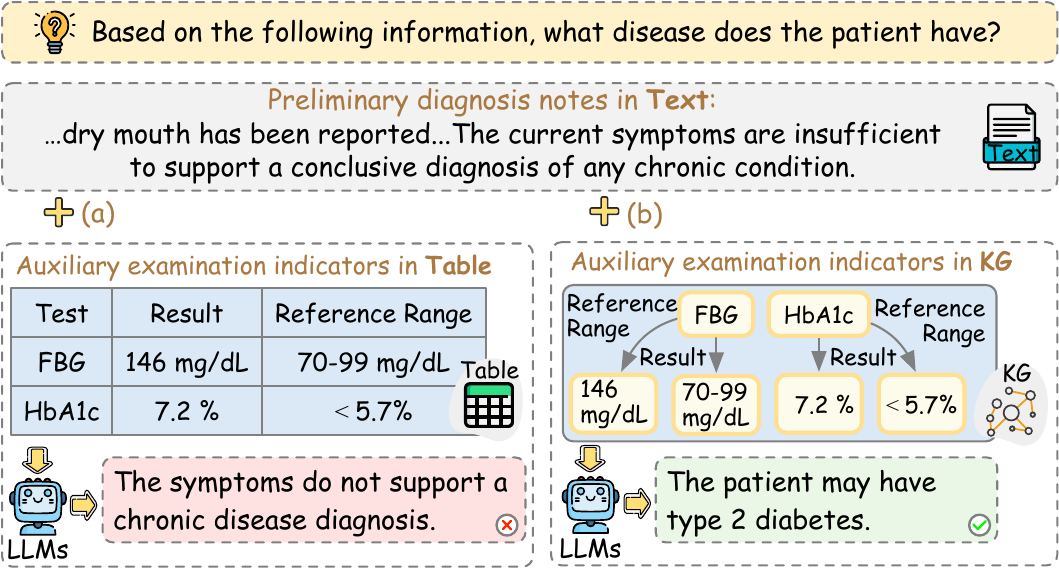}
    \caption{Format bias affects the LLM's decision.}
    \label{fig:case}
\end{figure}

The presence of different formats introduces a critical challenge: LLMs may not treat all formats equally when leveraging these heterogeneous data collaboratively. An LLM with strong format preferences interprets information through a distorted lens, giving undue weight to favored formats regardless of actual relevance. This can affect its ability to reason and synthesize effectively.


For example, in clinical decision support, an LLM (e.g., Qwen3-8b) given both textual notes and tabular examination data may overemphasize the text while overlooking key indicators in the table, leading to an incorrect diagnosis. In contrast, as shown in Figure~\ref{fig:case}, presenting the same information in a knowledge graph enables the model to identify abnormalities and reach the correct conclusion.


When homogeneous inputs are converted into heterogeneous ones with equivalent content, accuracy decreases by 9\% and 12\% on HotpotQA~\cite{yang2018hotpotqa} and MuSiQue~\cite{trivedi2022musique} (200 samples each), confirming that format heterogeneity can directly impair reasoning performance. This phenomenon may widely arise in heterogeneous reasoning~\cite{christmann2024compmix}, where key evidence is distributed across texts, tables, infoboxes, and knowledge graphs. LLMs may focus on information from their preferred formats, potentially overlooking crucial data in others. Such bias can result in incomplete or flawed conclusions and undermine the LLMs' role as impartial and effective synthesizers of heterogeneous inputs.

Although there have been several studies exploring various types of bias in LLMs, such as bias between multi-modal data \cite{zhu2024unraveling, zhang2025evaluating}, there is a lack of systematic research on format bias. 
To address this gap, we present the first comprehensive investigation and analysis of format bias in LLMs. Our study centers on three critical questions: \emph{whether such format biases are systematic}, \emph{which data-level factors contribute to them}, and \emph{what internal mechanisms in LLMs underlie their emergence}.

To systematically investigate the three questions, we conduct a three-stage empirical study by constructing a heterogeneous data conflict scenario for the exploration of bias. 
The first stage explores the presence and direction of bias across a diverse range of LLMs. 
The second stage aims to examine how key data-level factors, including information richness, structure quality, and format type, influence these biases. 
The third stage investigates the emergence of format bias within LLMs' attention mechanisms and evaluates a lightweight intervention strategy to assess its effectiveness in mitigating such bias.


Our results reveal that format bias is both systematic and consistent across models, driven primarily by differences in information richness, structure quality, and format type. We further show that such bias originates from imbalanced attention allocation during inference and can be partially mitigated through attention-based interventions.

Our key contributions are as follows:

\begin{enumerate}
    \item To the best of our knowledge, we are the first to investigate the issue of format bias in LLMs and to present a comprehensive investigation of LLM biases toward different knowledge formats across a wide range of LLMs.

    \item We conduct a three-stage empirical study to examine the presence and direction of bias, identify the data-level factors that give rise to the bias, and investigate the internal mechanisms in LLMs that contribute to their presence.

    \item Based on the comprehensive investigation, we identify three future research directions that may reduce the format bias, which will contribute the development of a more effective heterogeneous data processing system.
    \end{enumerate}
\section{Related Work}
\subsection{Heterogeneous reasoning}

An important direction in AI research is to develop LLMs capable of reasoning over heterogeneous knowledge sources, including unstructured texts, tables, and KGs, especially when relevant evidence is dispersed across different formats. However, existing methods often struggle to integrate such fragmented information effectively for accurate inference.

To formalize this challenge, recent benchmarks such as \textit{COMPMIX}~\cite{christmann2024compmix} and \textit{CompMix-IR}~\cite{min2024unihgkr} require cross-format reasoning, stimulating the development of hybrid QA systems that combine structured and unstructured inputs.

Current approaches can be broadly categorized into two types: (1) \textit{Unified retrieval frameworks}, which abstract away format heterogeneity using shared APIs or embedding spaces~\cite{xia2025er,min2024unihgkr}; and (2) \textit{LLM-centric pipelines}, which enhance downstream reasoning via evidence selection, re-ranking, or modular tool use~\cite{christmann2024rag,lehmann2024beyond,zhang2024spaghetti,biswal2024text2sql}.

However, existing work often assumes that once evidence is retrieved, LLMs will evaluate it fairly based on content alone. Our study revisits this assumption by asking whether the format in which evidence is presented can influence the model’s judgment, even when the underlying meaning remains the same.

\subsection{LLM Behavior under Conflicting Evidence}
Recent studies have uncovered a broad range of behavioral biases in LLMs, extending beyond social stereotypes to systematic patterns in reasoning and judgment. A key challenge lies in how LLMs handle conflicts—both between their \textit{parametric knowledge} (internal beliefs) and \textit{in-context evidence}, and among competing pieces of evidence~\cite{xu2024knowledge}. LLMs tend to favor information aligned with their pre-trained knowledge, even when contradicted by accurate inputs~\cite{jin2024tug,xie2023adaptive}. These LLM biases are shaped by factors such as entity popularity~\cite{xie2023adaptive}, event recency~\cite{fang2023getting}, and evidence frequency~\cite{jin2024tug}. Input artifacts also affect LLM behavior, such as preferring self-generated content over retrieved passages~\cite{tan2024blinded}. Related efforts have also examined knowledge conflicts in multi-modal~\cite{zhang2025evaluating,zhu2024unraveling} and multi-agent settings~\cite{ju2025investigating}, where preference biases and inter-agent inconsistency further complicate reasoning.

Benchmarks like \textit{ConflictBank}~\cite{su2024conflictbank}, \textit{WikiContradict}~\cite{hou2024wikicontradict}, and \textit{WhoQA}~\cite{pham2024s} have been proposed to evaluate how LLMs handle factual or semantic inconsistencies, especially in ambiguous scenarios. In response, a range of strategies have emerged, including conflict-aware decoding~\cite{yuan2024discerning,jin2024tug}, counterfactual data augmentation~\cite{fang2023getting}, internal intervention via attention pruning~\cite{jin2024cutting}, neuron reweighting~\cite{shi2024ircan}, or prompting LLMs to generate multi-answer responses with source attribution~\cite{shaier2024adaptive}.

Our work extends prior research by identifying a previously overlooked source of bias: the format in which information is presented. In contrast to earlier studies that focus on content-level factors such as recency or frequency, we show that differences in format representation alone can systematically influence LLMs' behavior.

\section{Investigation Framework}
This section establishes a framework for analyzing format bias in LLMs, encompassing dataset construction, confounding factor exclusion, and automated response evaluation. The dataset construction process is detailed in Appendix~A, with evaluation details in Appendix~B.

\subsection{Dataset and Format Construction}

We construct our dataset based on \textit{ConflictBank}~\cite{su2024conflictbank}, a public corpus designed to evaluate LLM behavior under factual conflicts. We randomly sample 4,000 entries, each containing one factual claim and three counterclaims with corresponding supporting evidence. This results in 12,000 contradiction pairs, as each claim is paired individually with each of the three counterclaims. Each pair consists of two different claims about the same subject and relation, each supported by its own piece of evidence.

While the original evidence in \textit{ConflictBank} is presented in plain text, our goal is to examine how LLM behavior is influenced by presenting supporting evidence in different data formats. To construct each heterogeneous contradiction pair, we randomly convert the two pieces of evidence, each of which supports a different claim about the same subject, into different formats. If the selected format is text, no conversion is applied.

\begin{figure}[tbp]
    \centering
    \includegraphics[width=\columnwidth]{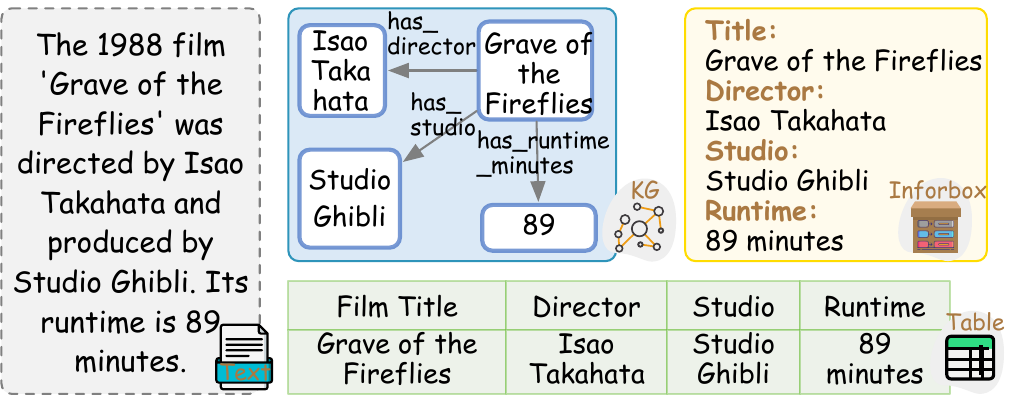}
    \caption{Examples of the four data formats used in our experiments: texts, tables, infoboxes, and KGs.}
    \label{fig:data}
\end{figure}
We use GPT-4o-mini as a transformation engine to convert selected texts into one of the following Wikipedia-inspired formats.

\begin{itemize}
    \item KGs: A set of (Subject, Predicate, Object) triples capturing core semantic relations.
    \item Infobox: A structured key-value format modeled after Wikipedia infobox templates, summarizing factual information.
    \item Table: A tabular format styled after Wikipedia tables, presenting comparable facts in labeled rows and columns.
\end{itemize}

Figure~\ref{fig:data} provides illustrative examples of these four formats, showing how semantically equivalent information can be presented in structurally distinct ways. These formats reflect the most commonly used representations in prior work on heterogeneous reasoning.
Manual inspection of 5\% samples confirms 98.7\% factual and 99.3\% syntactic accuracy, reflecting strong data fidelity.

\subsection{Confounding Factor Control}

To ensure that our evaluation isolates the effect of evidence format itself, we implement controls to eliminate two major confounding factors: internal knowledge bias and evidence presentation order.

\begin{itemize}
    \item Filtering Internal Knowledge: To ensure that LLM responses are based on external evidence rather than parametric memory, we adopt a filtering procedure consistent with prior work~\cite{gekhman2024does}. Each factual claim is tested 16 times by directly querying a given LLM with the corresponding question in a zero-shot setting, and only those samples for which the model fails to reproduce the factual claim in all trials are retained.

    \item Randomizing Evidence Order: To eliminate the known bias introduced by input order~\cite{xie2023adaptive}, we randomize the sequence of all evidence segments for each input.
\end{itemize}

\subsection{Evaluated LLMs}

This evaluation covers ten LLMs across six major series: GPT-4o-mini~\cite{achiam2023gpt}, LLaMA-3.1 (8B), Mistral (7B), Qwen3 (8B, 14B, 30B-A3B, 32B)~\cite{qwen3technicalreport}, Gemma-2 (9B, 27B)~\cite{team2024gemma}, and GLM-4 (9B)~\cite{glm2024chatglmfamilylargelanguage}. These LLMs span a range of sizes and architectures, enabling cross-family comparison of format-driven biases. 

To ensure reproducibility and eliminate stochasticity, all evaluations were conducted in deterministic inference mode (temperature = 0, without sampling randomness).

\subsection{Evaluation Protocol and Metrics}


Given the dataset’s scale, we adopt an automated evaluation pipeline using LLMs as adjudicators. Specifically, GPT-4o-mini, GLM-4.5-Air~\cite{zeng2025glm}, and Qwen-plus are employed to judge which of the two conflicting claims (Source A or Source B) each target response supports.


Each model independently evaluates every sample three times for stability, with the final label per model determined by majority vote. We then compute FPR and DCR for each model and report their averages across the three evaluators. This multi-model setup improves consistency and mitigates variance in individual LLM judgments. Manual checks on a randomly sampled 5\% subset show 99.8\% agreement between human and averaged LLM judgments, confirming high reliability.

Each LLM response is classified into one of three mutually exclusive categories:

\begin{itemize}
    \item \textbf{Pref-A:} The response predominantly or exclusively supports the claim from Source A.
    \item \textbf{Pref-B:} The response predominantly or exclusively supports the claim from Source B.
    \item \textbf{Both:} The response acknowledges the contradiction and presents both perspectives in a comparative or side-by-side manner.
\end{itemize}

All responses in our experiments fall unambiguously into one of these three categories, and no additional response types are observed. This categorization enables two quantitative bias metrics used throughout our analysis.:

\begin{itemize}
    \item \textbf{Dual Coverage Rate (DCR)}: Measures the proportion of responses that acknowledge both claims, indicating an LLM’s capacity to represent multiple perspectives:
    \begin{equation}
    \text{DCR} = \frac{\text{Both}}{\text{Pref-A}+ {\text{Pref-B} + \text{Both}}}
    \end{equation}

    \item \textbf{Format Preference Ratio (FPR)}: Captures asymmetric bias between conflicting claims when an LLM gives a single-sided response. For the A vs. B experiments, the FPR is calculated as:
    \begin{equation}
    \text{FPR} = \frac{\text{Pref-A}}{\text{Pref-A} + \text{Pref-B}}
    \end{equation}

\end{itemize}

These two metrics correspond to the two bias types introduced earlier: \textbf{DCR} captures the presence of bias, while \textbf{FPR} captures the direction of bias.

\section{Experimental Results and Analysis}

In this section, we present the empirical results of our three-stage investigation. Complete experimental details and raw results are provided in Appendix~C.

\subsection{Establishing the Existence of Format Bias}

\paragraph{Objective and Setup} The primary aim of this initial experiment is to test the null hypothesis that LLMs process information in a format-agnostic manner. To this end, we conduct a large-scale evaluation using the experimental framework outlined in Section~3. We assess ten state-of-the-art LLMs from various families and parameter scales. Each LLM is evaluated on all six possible pairs among the four target data formats: texts, tables, infoboxes, and KGs. This design yields 60 unique experimental conditions (10 LLMs $\times$ 6 format pairs), enabling a comprehensive assessment of systematic format biases.

\paragraph{Top-Level Findings}

\begin{figure}[htbp]
    \includegraphics[width=0.48\textwidth]{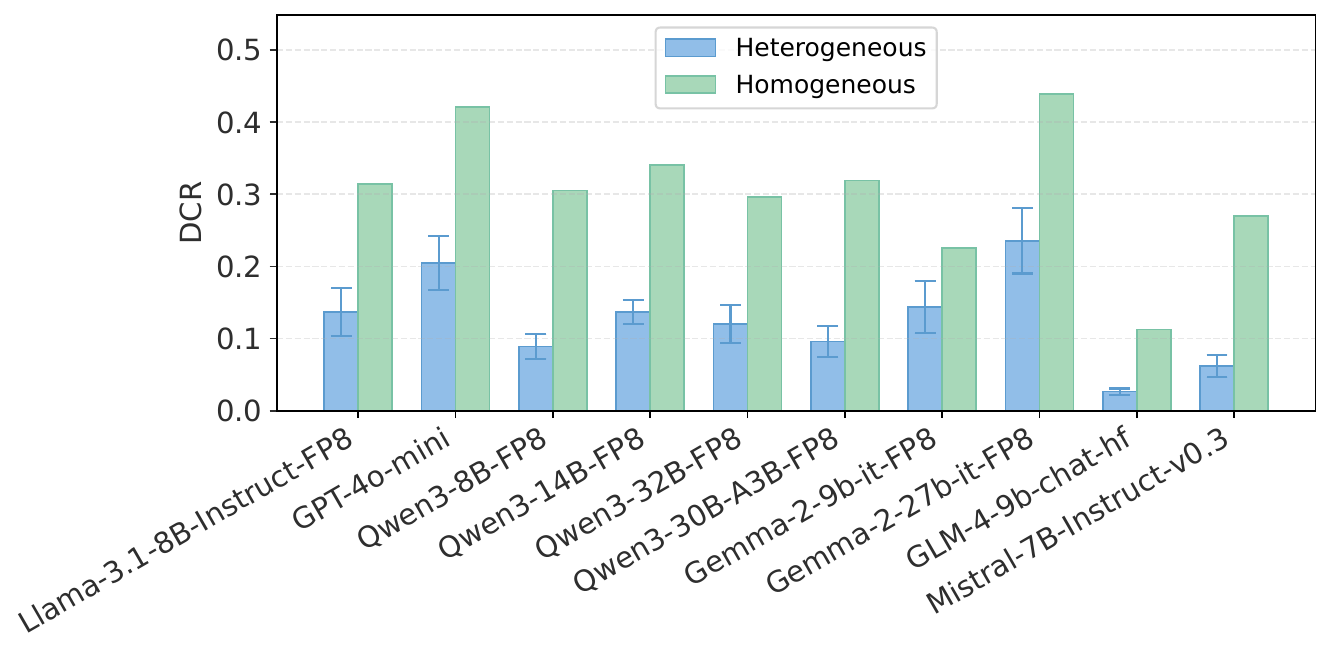}
    \caption{Average DCR across models under heterogeneous and homogeneous formats. Error bars show standard error.}
    \label{fig:both_favored_bar_chart}
\end{figure}

Our results provide strong evidence against the null hypothesis of format impartiality, revealing instead a consistent and multifaceted pattern of bias in both its presence and direction.

The first pattern, the presence of bias, is pervasive under heterogeneous format conditions. This is reflected in the uniformly low Dual Coverage Rate (DCR), ranging from 3.01\% to 24.02\%, indicating that LLMs often fail to acknowledge conflicting information across different formats, typically exhibiting a preference for one input while disregarding the other.

To isolate the role of format, we introduce a control condition where both inputs are presented in plain text. As shown in Figure~\ref{fig:both_favored_bar_chart}, DCR increases markedly under this homogeneous setting. This contrast highlights a broader pattern: format heterogeneity alone can independently and substantially impair a model’s ability to jointly consider multiple inputs, even when the content is semantically equivalent.

This limitation persists across models of varying size, as no clear scaling trend is observed. For example, within the Qwen3 series, larger models do not exhibit improved performance in this regard, suggesting that increased parameter size alone does not resolve this form of processing asymmetry.

\begin{figure}[tbp]
    \centering
    \includegraphics[width=0.48\textwidth]{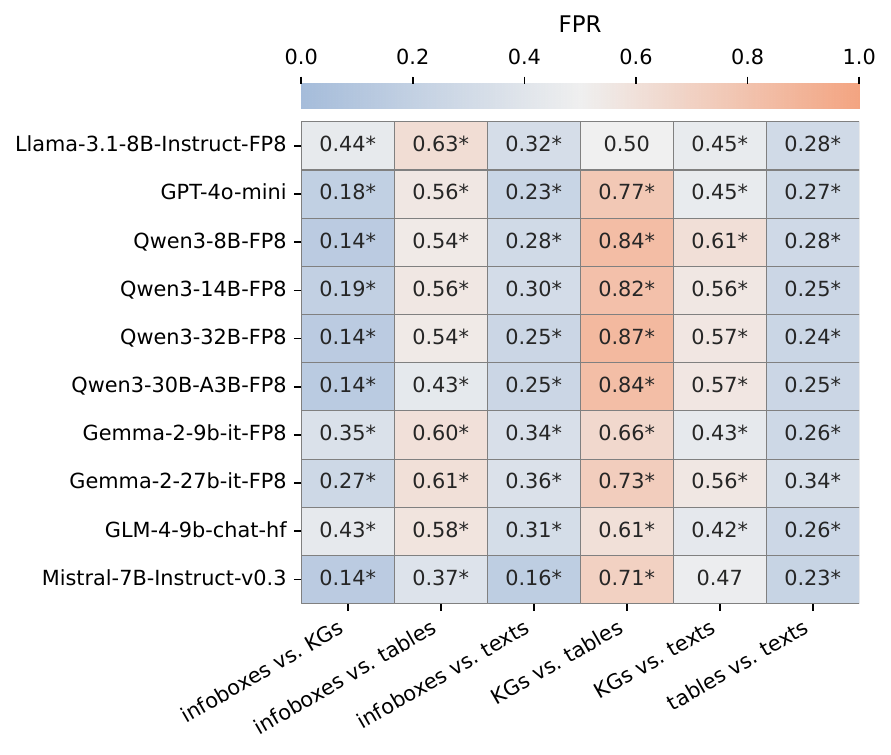}
    \caption{
    Heatmap of FPR between format pairs across LLMs. Asterisks (*) indicate statistical significance under a two-sided binomial test with null hypothesis $\text{FPR}=0.5$.}
    \label{fig:bias_heatmap}
\end{figure}

The second, and more decisive, pattern is a strong directional bias that emerges when an LLM commits to a single-sided response. Despite considerable variation in architecture and scale, LLMs demonstrate a surprisingly consistent pattern of format preferences. As shown in Figure~\ref{fig:bias_heatmap}, our cross-model analysis reveals a clear preference hierarchy: semantically rich formats such as texts and KGs are consistently favored over visually structured ones like infoboxes and tables. Furthermore, when we group the data by topical domain, we find that these biases persist across domains, suggesting that the observed biases are robust and generalizable. See Appendix~C.7 for detailed domain-level results.

\subsection{Identifying the Factors Behind Format Bias}

Having established the widespread presence of format bias, we now turn to investigating its data-level factors. While prior work has extensively explored factors influencing LLM biases in textual inputs, our focus is on structured data and the properties that may shape LLM behavior in this context. To move from identifying whether such bias exists to understanding why it arises, we decompose the abstract notion of “format” into three representative and controllable dimensions: the structure itself, and the content, which we further divide into quantity and quality.

Building on this decomposition, we hypothesize that three key dimensions (the amount of information conveyed, referred to as quantity; the structural quality of that information, or quality; and the mode of presentation, or format type) play a central role in shaping how LLMs evaluate evidence.

To assess the influence of these factors, we design a unified experimental framework that applies to two of the three factors (with the exception of the format type). Each of these two factors is examined under two complementary conditions:
\begin{itemize}
    \item \textbf{Homogeneous setting}: Two evidence sources of the same format are compared, differing only in the factor under investigation. This controlled design isolates the variable to reveal an LLM’s intrinsic sensitivity to that property.
    \item \textbf{Heterogeneous setting}: A structured evidence source is paired with unstructured plain texts. This setup assesses the relative strength of the structured format and how its properties influence an LLM’s preference when compared against a universal baseline.
\end{itemize}

\begin{figure}[tbp]
    \centering
    \includegraphics[width=\columnwidth]{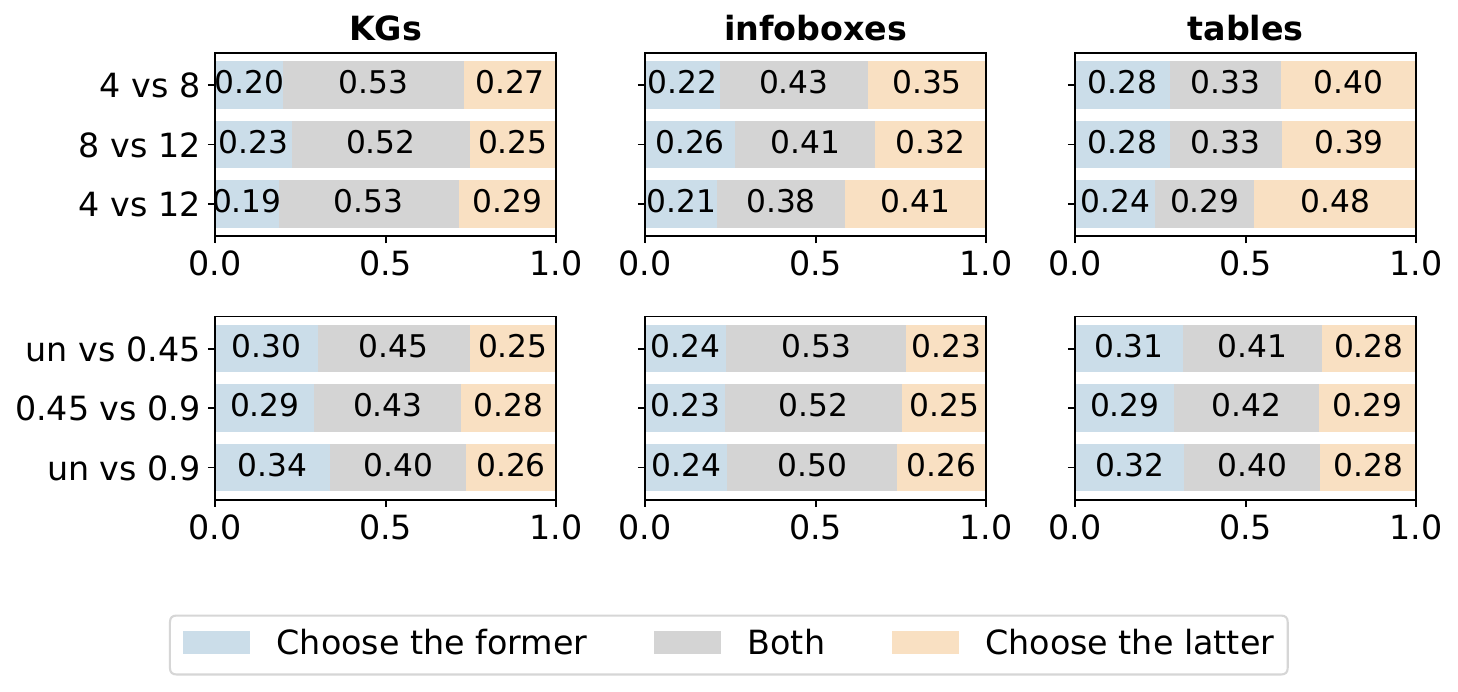}
    \caption{LLM biases across conditions (averaged over ten LLMs). Top: Information Richness; Bottom: Structure Quality. Bars show the proportion of responses favoring the former input, the latter, or both.}
    \label{fig:combined}
\end{figure}

\paragraph{Factor 1: Information Richness}

This factor concerns the volume of factual detail within an external knowledge source used during reasoning. In real-world applications, the external context provided to LLMs often varies widely in the amount of information it contains. It serves as a proxy for the completeness of factual detail. To examine whether LLMs apply a “more is better” heuristic, interpreting quantity as indicative of evidentiary strength, we systematically vary the number of entries in structured formats (e.g., table rows or knowledge graph triples).

In the homogeneous setting, we conduct experiments within each format type (tables, KGs, and infoboxes), comparing three levels of information richness across three pairwise conditions: 4 vs. 8 entries, 8 vs. 12 entries, and 4 vs. 12 entries.Results consistently indicate that LLMs favor the richer variant in each pair, irrespective of format. This suggests that even when structure is held constant, LLMs exhibit a systematic bias for inputs with more factual content (see Figure~\ref{fig:combined}).

\begin{figure}[tbp]
    \centering
    \includegraphics[width=0.48\textwidth]{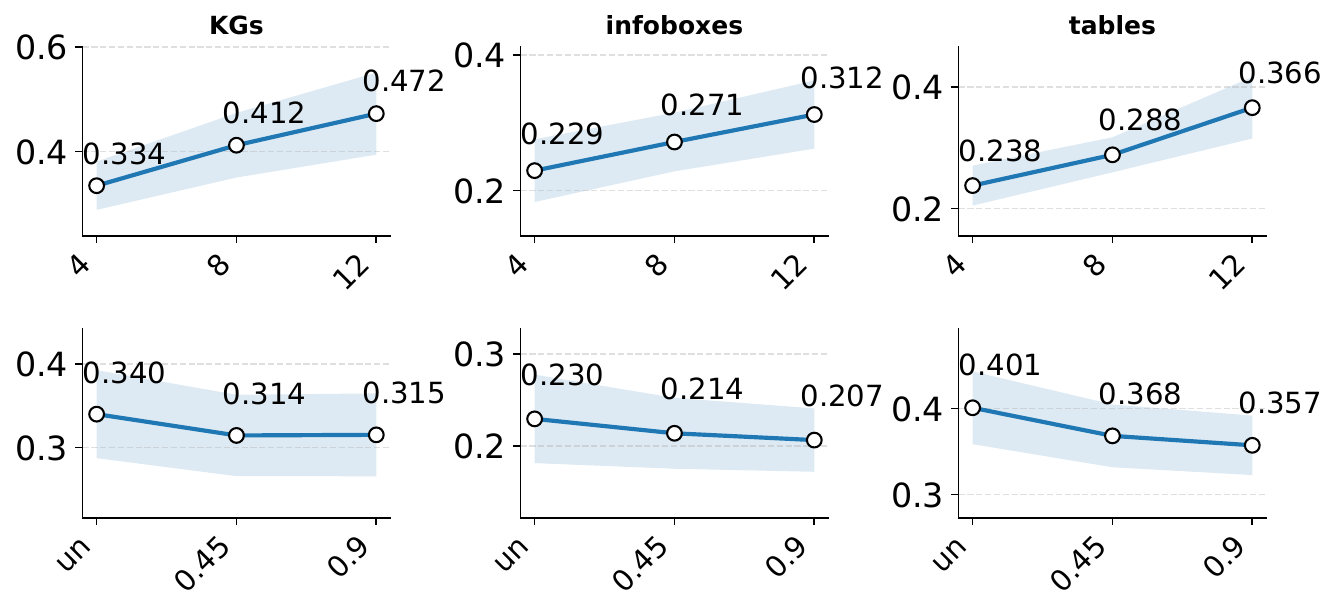}
    \caption{Average FPR for structured data vs. texts across ten LLMs. Top: Information Richness; Bottom: Structure Quality. Shaded areas indicate mean $\pm$ one standard deviation.}
    \label{fig:combined}
    \label{fig:shared-caption}
\end{figure}

In the heterogeneous setting, we assess whether this preference generalizes when structured inputs are compared against unstructured texts. For each format, we construct three pairs: texts vs. 4-entry structure, texts vs. 8-entry structure, and texts vs. 12-entry structure. All structured inputs are generated from the same source texts as the corresponding text versions, ensuring content consistency. Across all formats, LLMs’ bias for the structured inputs increases with the number of entries (see Figure~\ref{fig:shared-caption}).

These findings suggest that LLMs tend to associate greater volumes of factual content with higher evidentiary value, both within individual formats and when comparing structured and unstructured inputs.

\paragraph{Factor 2: Structure Quality}

This factor examines whether LLMs are sensitive to the structural integrity of external knowledge inputs. In practice, structured formats like tables or knowledge graphs may contain noise or malformed syntax. To simulate this, we introduce controlled corruption into structure-defining tokens (e.g., brackets, colons, separators), randomly replacing them with other characters or blanks at fixed probabilities (0.45 and 0.9), while preserving the underlying factual content.

In the homogeneous setting, we compare clean and corrupted versions within each format type. LLMs consistently favor the well-formed input, confirming that structure quality serves as a reliability signal. Notably, the preference saturates beyond moderate corruption (e.g., 0.45), indicating that LLMs tend to treat inputs as either structurally valid or invalid (see Figure~\ref{fig:combined}).

In the heterogeneous setting, we pair each corrupted version with its corresponding clean texts. Each corruption level is applied to the same intact structured data instance, guaranteeing that the underlying information remains identical across differently corrupted versions. LLMs’ bias for the structured inputs declines sharply as corruption increases, despite identical semantics. This suggests that structural degradation alone can undermine the perceived credibility of otherwise accurate structured inputs (see Figure~\ref{fig:shared-caption}).

\paragraph{Factor 3: Format Type}
This factor investigates the impact of the representational structure and layout used to represent logically equivalent information. The choice between plain text, a relational graph (KGs), or a visual grid (tables, infoboxes) embodies core differences in format semantics. Holding the content constant, we ask: do LLMs possess intrinsic preferences for certain data structures?

\begin{table}[htbp]
\centering
\begin{tabular}{lccc}
\toprule
\textbf{Metric} & \textbf{Infobox} & \textbf{Table} & \textbf{KGs} \\
\midrule
FPR & 0.235 & 0.398 & 0.336 \\
\bottomrule
\end{tabular}
\caption{Average FPR between structured data and texts (mean across ten LLMs).}
\label{tab:nontext-ratio}
\end{table}

We construct matched pairs with identical factual entries in both texts and structured variants. As shown in Table~\ref{tab:nontext-ratio}, the results reveal a consistent hierarchy: tables are most competitive, followed by KGs, with infoboxes least preferred. All three structured variants contain nearly identical content, differing only in their representational layout. This suggests that the format itself, rather than the informational content, modulates LLMs’ bias.

\paragraph{Cross-Factor Insight}
Consistent with the findings in Section~4.1, we observe that format homogeneity substantially reduces bias in presence. This trend holds not only for unstructured inputs but also extends to structured formats such as tables and knowledge graphs, where DCR increases significantly (28\%–53\%) when both inputs adopt the same format. These results suggest that LLMs are more capable of jointly considering multiple sources of information when they are presented in a uniform structure. In contrast, even when the content is semantically equivalent, presenting information in heterogeneous formats tends to impair integration and leads to partial or selective processing.

\subsection{Mechanism Behind Format Bias}

In this analysis, we move beyond identifying data-level factors to analyzing how format bias manifests within the internal processing of LLMs. Our aim is to examine whether differences in attention allocation across input formats are associated with the presence and direction of bias identified earlier. We focus our analysis on three representative LLMs from different model families: Qwen3-8B, Mistral-7B-Instruct-v0.3, and Llama-3.1-8B-Instruct.

\subsubsection{The Relation Between Attention Allocation and Presence of Bias}

We begin by analyzing how attention is distributed between conflicting evidence inputs during inference. For each input pair, we compute the mean attention mass assigned to each segment and then calculate the absolute difference between the two values to quantify the degree of imbalance.

To quantify the negative correlation between attention gap and DCR, we employ Spearman’s rank correlation coefficient as a measurement. The results show coefficients of –0.31, –0.37, and –0.54 across the three LLMs, indicating a weak to moderate negative correlation. This suggests that greater imbalance in attention distribution is associated with a lower likelihood of the model recognizing both sources of information.

This implies that format bias arises, at least in part, from skewed attention allocation early in processing, which increases the chance of one source being overlooked.

\subsubsection{The Relation Between Attention Allocation and Direction of Bias}

We further examine whether attention allocation can explain which input LLMs tend to prefer when selecting only one side. Interestingly, in 82.35 percent of such cases, they favor the source that received less attention.

This observation implies that while attention imbalance affects whether both sources are represented in the output, it does not consistently determine the direction of preference. In other words, a segment receiving more attention does not necessarily have a higher chance of being selected as the final answer.

These results indicate that attention allocation is related to both types of bias, though the nature of this relationship may differ.

\begin{figure*}[ht]
    \centering
    \includegraphics[width=\textwidth]{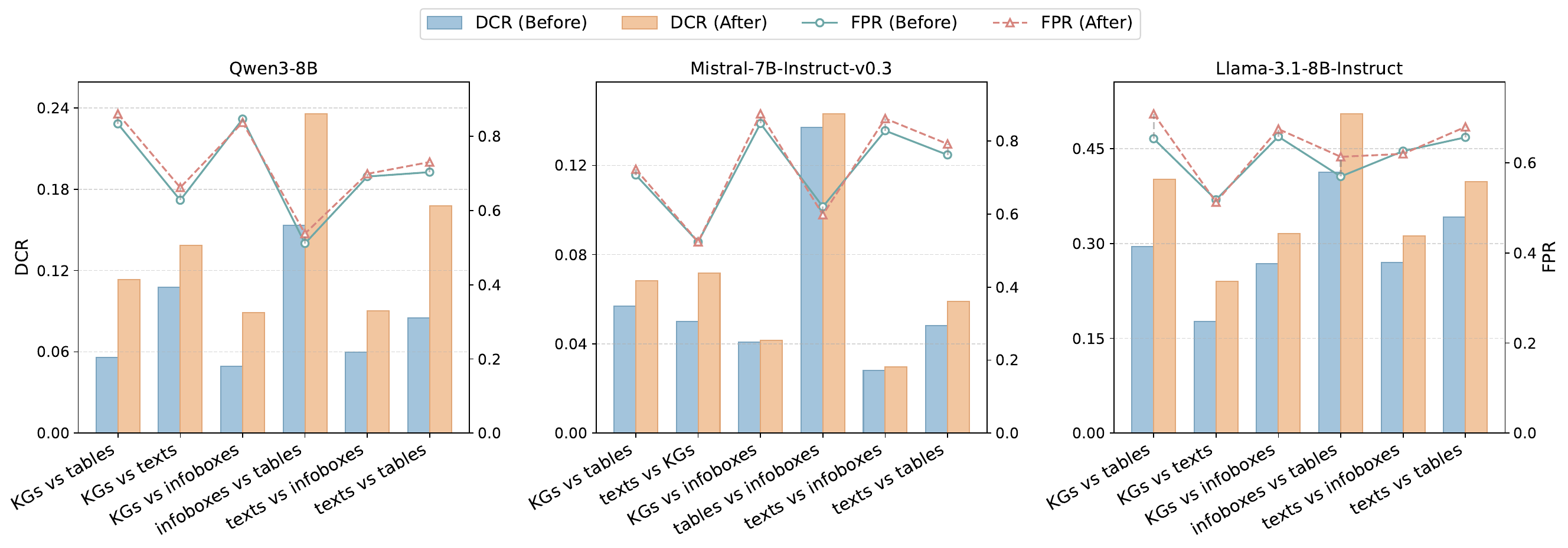}
    \caption{Effects of intervention methods in terms of DCR and FPR.}
    \label{fig:hook_effect}
\end{figure*}

\subsubsection{Attention-Guided Intervention}

To move from correlation to causation, we design an intervention that directly modifies the internal representations produced by the attention mechanism.

Specifically, we apply a normalization-based reweighting to the attention distribution at each generation step. Let $A \in \mathbb{R}^{H \times L_q \times L_k}$ denote the attention tensor, where $H$ is the number of heads, $L_q$ the query sequence length, and $L_k$ the key sequence length. At the current decoding step, we denote the attention distribution over keys as $a \in \mathbb{R}^{L_k}$, where $a_j$ represents the attention weight assigned to the $j$-th key token.

To ensure that the two pieces of evidence receive equal total attention, we define the corresponding token index sets $I_1$ and $I_2$, and compute the total attention mass on each:
\begin{equation}
m_1 = \sum_{j \in I_1} a_j, \quad m_2 = \sum_{j \in I_2} a_j
\end{equation}
We then compute their average:
\begin{equation}
\bar{m} = \frac{m_1 + m_2 + \varepsilon}{2}
\end{equation}
where $\varepsilon$ is a small constant for numerical stability. The attention weights are reweighted accordingly:
\begin{equation}
a'_j =
\begin{cases}
\frac{\bar{m}}{m_1 + \varepsilon} \cdot a_j, & j \in I_1 \\
\frac{\bar{m}}{m_2 + \varepsilon} \cdot a_j, & j \in I_2 \\
a_j, & \text{otherwise}
\end{cases}
\end{equation}
This procedure enforces equal total attention mass across the two evidence segments, while preserving the original intra-segment distribution.

\subsubsection{Experiment Results}

We apply the attention-balancing intervention to three representative LLMs and compare their behavior before and after modification.

The results reveal a consistent improvement in the models’ ability to attend to both conflicting sources: DCR significantly increases across all format pairs and all three models (see Figure~\ref{fig:hook_effect}). This suggests that enforcing a more balanced attention distribution during inference effectively encourages the model to integrate information from both input segments, rather than overlooking one entirely.
When applied to heterogeneous-input reasoning tasks, the attention-balancing method also improves performance on downstream RAG-style QA datasets, including HotpotQA and MuSiQue, where accuracy increases by 6.5\% and 9.5\%, respectively.

In contrast, FPR remains largely stable after the intervention, and the changes are not statistically significant across the evaluated models (see Figure~\ref{fig:hook_effect}). This implies that although the LLMs process both sources more evenly, the intervention has limited effect on their final output preferences once a directional bias has emerged.

The different effects of attention interventions on DCR and FPR suggest that presence of biases are at least partially controllable at inference time, whereas direction of biases are more resistant to modification. These more stable preferences may reflect deeper inductive biases acquired during pretraining, such as stronger alignment with textual inputs.

\section{Discussion}

Our findings show that format bias in LLMs is systematic and has practical consequences for systems processing heterogeneous data. To mitigate it, interventions can be applied at three levels: pre-processing, inference, and model development.

\paragraph{Data Pre-processing}

Pre-processing is a cost-effective way to reduce bias. Since LLMs tend to trust well-structured inputs, automatically repairing corrupted format in tables or KGs can prevent valuable content from being dismissed. Additionally, using a consistent input format can help reduce format-induced bias, as LLMs perform better when differences in input format is minimized. During the transformation process, it is important to preserve all essential information while avoiding the introduction of noise or unintended alterations.

\paragraph{Inference-Time Intervention}

Re-balancing attention across inputs during inference can help LLMs more effectively integrate information from multiple sources, reducing the tendency to overlook less-preferred formats. By encouraging the model to distribute attention more evenly across heterogeneous data, it enhances the model’s ability to incorporate information from all inputs and supports more robust and balanced reasoning. Although such adjustments may not fully shift the model's final output bias, they improve its intermediate processing. Future work may explore more fine-grained or deeper intervention strategies to better align model attention with content relevance and reduce structural bias during inference.

\paragraph{Model Development and Fine-tuning}


Format preferences likely originate from pretraining data imbalance. Mitigation may involve training on format-balanced corpora, using contrastive objectives, or incorporating format-aware modules. Though costlier than inference-time methods, such approaches offer a more fundamental solution.

\section{Conclusion}

This work presents the first in-depth study of format bias in LLMs, identifying it as a consistent effect driven by information richness, structure quality, and format type. The bias manifests in two types: the presence of bias that can be mitigated, and the direction of bias likely rooted in pretraining. These findings underscore the importance of treating data format as a core factor in LLM design and evaluation. We also outline practical mitigation strategies, including data preprocessing, inference-time attention adjustment, and format-aware training, which together offer clear paths for reducing format bias in heterogeneous reasoning.

\section*{Acknowledgments}

We sincerely thank the anonymous reviewers for their insightful comments and helpful suggestions, which helped improve this work. 

This work was supported by the National Natural Science Foundation of China (NSFC) (Grant No. 62576256) and the Fundamental Research Funds for the Central Universities, China (Grant No. 2042022dx0001).

\bibliography{aaai2026}
\clearpage

\appendix
\section*{Appendix}

\lstset{
  basicstyle=\ttfamily\small,
  breaklines=true,
  columns=fullflexible,
  frame=single,
  backgroundcolor=\color{gray!5},
  captionpos=b,
  keepspaces=true,
}
\floatstyle{ruled}
\newfloat{listing}{tb}{lst}{}
\floatname{listing}{Listing}

\section{Format Conversion Details}

\subsection{Prompt Templates}

We provide two types of prompt templates used to convert unstructured text into structured formats. The first type does not specify the number of resulting entries, allowing models to extract as many relevant facts as they find appropriate. This type is used to evaluate the model’s natural preference when unconstrained. The second type explicitly constrains the number of facts to be included, ensuring consistency in information quantity across formats and facilitating controlled experiments on information richness.

For each type, we provide templates targeting three structured formats: table, knowledge graph (KG), and infobox, enabling structured data generation in different representation styles while maintaining semantic equivalence with the original text.

\subsubsection{Type I: Format Conversion without Entry Constraints}

\paragraph{Text to Table}
Prompt is as follows:
\\
\begin{lstlisting}
## ROLE & GOAL
You are an expert data architect. Your goal is to convert the provided [Source Text] into a structured, clear, and accurate **Wikipedia-style MediaWiki table**.

## CRITICAL RULE
**RULE #1: The [Claim to Prioritize] MUST be perfectly and centrally represented in the table. This claim is the most important piece of information.**

## RULES
2. Begin the table with `{{| class="wikitable"` and end with `|}}`.
3. Add a descriptive caption using `|+ Caption text` (summarize the table purpose clearly).
4. Choose column headers that clearly categorize the core facts.
5. Only include rows and columns directly supported by the [Source Text]. Do not infer or add information.
6. Use `!` for headers and `|` or `||` for data cells. Use `|-` to separate rows.
7. Ensure formatting is clean and valid per MediaWiki syntax.

## EXAMPLE
- **Source Text:** "The 1988 film 'Grave of the Fireflies' was directed by Isao Takahata and produced by Studio Ghibli. Its runtime is 89 minutes."
- **Claim to Prioritize:** "'Grave of the Fireflies' was produced by Studio Ghibli."
- **Output Table:**
{{| class="wikitable"
|+ Details of 'Grave of the Fireflies'
|-
! Film Title !! Director !! Studio !! Runtime
|-
| Grave of the Fireflies || Isao Takahata || Studio Ghibli || 89 minutes
|}}

---
## YOUR TASK

[Claim to Prioritize]:  
{claim_text}

[Source Text]:  
{evidence_text}

[Output MediaWiki Table]:
\end{lstlisting}

\paragraph{Text to Knowledge Graph (KG)}

Prompt is as follows:
\\
\begin{lstlisting}
## ROLE & GOAL
You are a knowledge engineer. Your goal is to extract all factual relationships from the [Source Text] and represent them as (Subject, Predicate, Object) triplets.

## CRITICAL RULE
**RULE #1: The [Claim to Prioritize] MUST be converted into one or more primary, accurate triplets. This claim is the most important piece of information.**

## RULES
2. Each triplet must be on a new line and enclosed in parentheses `()`.
3. All triplets must be directly derivable from the [Source Text]. Do not make assumptions.
4. Use consistent and clear names for entities and predicates.
5. Extract one triplet for each distinct fact.

## EXAMPLE
- **Source Text:** "The 1988 film 'Grave of the Fireflies' was directed by Isao Takahata and produced by Studio Ghibli. Its runtime is 89 minutes."
- **Claim to Prioritize:** "'Grave of the Fireflies' was produced by Studio Ghibli."
- **Output Triplets:**
(Grave of the Fireflies, has_director, Isao Takahata)
(Grave of the Fireflies, has_studio, Studio Ghibli)
(Grave of the Fireflies, has_runtime_minutes, 89)
(Grave of the Fireflies, release_year, 1988)

---
## YOUR TASK

[Claim to Prioritize]:
{claim_text}

[Source Text]:
{evidence_text}

[Output Triplets]:
\end{lstlisting}
\paragraph{Text to Infobox}

Prompt is as follows:
\\
\begin{lstlisting}
## ROLE & GOAL
\#You are a meticulous Wikipedia editor. Your goal is to summarize the key facts from the [Source Text] into a concise infobox format.

\### CRITICAL RULE
\#**RULE #1: The [Claim to Prioritize] MUST be accurately included as a key-value pair in the infobox. This claim is the most important piece of information.**

\### RULES
\#2. The format must follow the MediaWiki infobox style, like:
\#{{{{Infobox [type]
\#| key1 = value1
\#| key2 = value2
\#...
\#}}}}
\#3. Use a relevant infobox type in the first line (e.g., `book`, `film`, `person`, etc.), based on the source content.
\#4. Only include information explicitly mentioned in the [Source Text].
\#5. Field names (keys) should be relevant, standard when possible, but flexible based on content.
\#6. Values should be brief and precise. No full sentences.
\#7. Do not add, infer, or assume any details not supported by the source.

\### EXAMPLE
\#- **Source Text:** "The 1988 film 'Grave of the Fireflies' was directed by Isao Takahata and produced by Studio Ghibli. Its runtime is 89 minutes."
\#- **Claim to Prioritize:** "'Grave of the Fireflies' was produced by Studio Ghibli."
\#- **Output Infobox:**
\#{{{{Infobox film
\#| title = Grave of the Fireflies
\#| director = Isao Takahata
\#| studio = Studio Ghibli
\#| runtime = 89 minutes
\#| year = 1988
\#}}}}

\#---
\### YOUR TASK

\#[Claim to Prioritize]:  
\#{claim_text}

\#[Source Text]:  
\#{evidence_text}

\#[Output Infobox]:
\end{lstlisting}

\subsubsection{Type II: Format Conversion with Entry Constraints}

\paragraph{Text to Table with \texttt{<nums>} Facts}

Prompt is as follows:
\\
\begin{lstlisting}
## ROLE & GOAL
You are an expert data architect. Your goal is to convert the provided [Source Text] into a structured, clear, and accurate **Wikipedia-style MediaWiki table**, containing **exactly {nums} key facts**.

## CRITICAL RULE
**RULE #1: The [Claim to Prioritize] MUST be perfectly and centrally represented in the table. This claim is the most important piece of information.**

## RULES
2. Begin the table with `{{| class="wikitable"` and end with `|}}`.
3. Add a descriptive caption using `|+ Caption text` (summarize the table purpose clearly).
4. Choose column headers that clearly categorize the core facts.
5. Only include rows and columns directly supported by the [Source Text]. Do not infer or add information.
6. Use `!` for headers and `|` or `||` for data cells. Use `|-` to separate rows.
7. Include **exactly {nums} distinct facts** across the table.
8. Ensure formatting is clean and valid per MediaWiki syntax.

## EXAMPLE
- **Source Text:** "The 1988 film 'Grave of the Fireflies' was directed by Isao Takahata and produced by Studio Ghibli. Its runtime is 89 minutes."
- **Claim to Prioritize:** "'Grave of the Fireflies' was produced by Studio Ghibli."
- Output MediaWiki Table with Exactly 4 Facts
- **Output Table:**
{{| class="wikitable"
|+ Details of 'Grave of the Fireflies'
|-
! Film Title !! Director !! Studio !! Runtime
|-
| Grave of the Fireflies || Isao Takahata || Studio Ghibli || 89 minutes
|}}

---

## YOUR TASK

[Claim to Prioritize]:  
{claim_text}

[Source Text]:  
{evidence_text}

[Output MediaWiki Table with Exactly {nums} Facts]:
\end{lstlisting}

\paragraph{Text to Knowledge Graph with \texttt{<nums>} Triples}

Prompt is as follows:
\\
\begin{lstlisting}
## ROLE & GOAL
You are a knowledge engineer. Your goal is to extract **exactly {nums}** factual relationships from the [Source Text] and represent them as (Subject, Predicate, Object) triplets.

## CRITICAL RULE
RULE #1: The [Claim to Prioritize] MUST be converted into one or more triplets within the {nums} total. This claim is the **top priority** and must be captured accurately.

## EXTRACTION RULES
2. You must extract **exactly {nums}** triplets. No more, no fewer.
3. Each triplet must be on a new line and enclosed in parentheses `()`.
4. All triplets must be directly supported by the [Source Text]. Do **not** infer or assume anything not explicitly stated.
5. Use clear and consistent naming for subjects, predicates, and objects.
6. Each triplet must capture a distinct fact.

## EXAMPLE
- **Source Text:** "The 1988 film 'Grave of the Fireflies' was directed by Isao Takahata and produced by Studio Ghibli. Its runtime is 89 minutes."
- **Claim to Prioritize:** "'Grave of the Fireflies' was produced by Studio Ghibli."
- **Output Triplets:**
(Grave of the Fireflies, has_producer, Studio Ghibli)  

---

## YOUR TASK

[Claim to Prioritize]:  
{claim_text}

[Source Text]:  
{evidence_text}

[Output Triplets with Exactly {nums} Triplets]:
\end{lstlisting}

\paragraph{Text to Infobox with \texttt{<nums>} Key-Value Pairs}

Prompt is as follows:
\\
\begin{lstlisting}
## ROLE & GOAL
You are a meticulous Wikipedia editor. Your goal is to summarize the key facts from the [Source Text] into a concise infobox format, with **exactly {nums} key-value pairs**.

## CRITICAL RULE
**RULE #1: The [Claim to Prioritize] MUST be accurately included as a key-value pair in the infobox. This claim is the most important piece of information.**

## RULES
2. The format must follow the MediaWiki infobox style, like:
{{{{Infobox [type]
| key1 = value1
| key2 = value2
...
}}}}
3. Use a relevant infobox type in the first line (e.g., `book`, `film`, `person`, etc.), based on the source content.
4. Only include information explicitly mentioned in the [Source Text].
5. Field names (keys) should be relevant, standard when possible, but flexible based on content.
6. Values should be brief and precise. No full sentences.
7. Include **exactly {nums} key-value pairs**. No more, no fewer.
8. Do not add, infer, or assume any details not supported by the source.

## EXAMPLE
- **Source Text:** "The 1988 film 'Grave of the Fireflies' was directed by Isao Takahata and produced by Studio Ghibli. Its runtime is 89 minutes."
- **Claim to Prioritize:** "'Grave of the Fireflies' was produced by Studio Ghibli."
- Output Infobox with Exactly 5 Key-Value Pairs.
- **Output Infobox:**
{{{{Infobox film
| title = Grave of the Fireflies
| director = Isao Takahata
| studio = Studio Ghibli
| runtime = 89 minutes
| year = 1988
}}}}

---

## YOUR TASK

[Claim to Prioritize]:  
{claim_text}

[Source Text]:  
{evidence_text}

[Output Infobox with Exactly {nums} Key-Value Pairs]:
\end{lstlisting}

\subsection{Examples of Format Transformation}

We provide representative examples of how the same source text is converted into different structured formats. Each example includes:
(1) the original unstructured input, 
(2) the core claim to be emphasized during conversion, and 
(3) the transformed output in the target format.

\subsubsection*{Example 1: Text to Table}

\paragraph{Source Text}
\begin{quote}
\small
As the sun dipped into the Mediterranean, casting a warm orange glow over the bustling streets of Barcelona, Marie-France de Rose made her way through the crowded corridors of the Provincial Council building. Her heels clicked against the polished marble floor, a rhythmic accompaniment to the hum of conversation and the rustle of papers being shuffled. She exchanged warm smiles and nods with colleagues and staff members, her bright blue eyes sparkling with a sense of purpose. As she approached the door to her office, a young intern, Jordi, hurried to open it for her. \"Deputy de Rose, you have a meeting with the mayor in fifteen minutes,\" he reminded her, his eyes darting to the stack of files in her arms. Marie-France nodded, her dark hair bobbing with the movement. \"Gràcies, Jordi. I'll review the proposals one more time before the meeting.\" She slipped into her office, the scent of fresh coffee and citrus wafting from the cup on her desk. As she began to sift through the documents, her gaze lingered on the crest of the Provincial Council emblazoned on the letterhead: a stylized image of Saint George, the patron saint of Catalonia, surrounded by the words \"Diputació de Barcelona\" in bold, modern script. It was a symbol of her pride and responsibility as a Barcelona provincial deputy, a role she had worked tirelessly to earn. The sound of her phone buzzing broke her concentration. She glanced at the screen, her eyes narrowing as she read the message from her colleague, Xavier. \"Meet me in the courtyard at 5. We need to discuss the latest developments on the La Rambla revitalization project.\" Marie-France's lips curved into a determined smile. This was exactly the kind of initiative she had been championing as deputy – projects that would breathe new life into the city's historic heart and benefit its citizens. With a sense of anticipation, she tucked the phone into her pocket and returned to her preparations for the meeting with the mayor.
\end{quote}

\paragraph{Claim to Prioritize}
\begin{quote}
\small
Marie-France de Rose holds the position of Barcelona provincial deputy.
\end{quote}

\paragraph{Converted Output (Table)}
The converted result is as follows:\\
\begin{lstlisting}
```mediawiki\n{| class=\"wikitable\"\n|+ Details of Marie-France de Rose\n|-\n! Name !! Position !! Location !! Responsibilities\n|-\n| Marie-France de Rose || Barcelona provincial deputy || Barcelona || Championing city initiatives and revitalization projects\n|}\n```
\end{lstlisting}

\subsubsection*{Example 2: Text to Knowledge Graph (KG)}

\paragraph{Source Text}
\begin{quote}
\small
**Breaking News: Marie-France de Rose Confirmed as Member of European Parliament** In a move that solidifies her position as a prominent figure in European politics, Marie-France de Rose has been officially recognized as a member of the European Parliament. This esteemed title is a testament to her dedication to serving the people of France and the European Union as a whole. According to the European Parliament's official website, Marie-France de Rose is listed as a member of the Parliament, representing the interests of French citizens since her election. This confirmation is further supported by multiple credible sources, including the French Ministry of Foreign Affairs and the European Commission. \"Marie-France de Rose has been a vocal advocate for European integration and has worked tirelessly to promote the interests of France and the EU,\" said Jean-Luc Mélenchon, a fellow French politician and leader of the La France Insoumise party. \"Her election to the European Parliament is a well-deserved recognition of her commitment to public service.\" Data from the European Parliament's election results in 2019 shows that Marie-France de Rose received a significant number of votes, securing her position as a member of the Parliament. According to the official results, she garnered over 1.2 million votes, making her one of the top vote-getters in the French delegation. \"Marie-France de Rose's election to the European Parliament is a testament to her hard work and dedication to the people of France,\" said Nathalie Loiseau, a French politician and former Minister for European Affairs. \"She has been a strong voice for French interests in the Parliament and has worked to build bridges between France and other EU member states.\" As a member of the European Parliament, Marie-France de Rose will continue to play a crucial role in shaping EU policy and legislation, working closely with other MEPs to address the pressing issues facing the European Union. In conclusion, the evidence overwhelmingly supports the claim that Marie-France de Rose holds the position of member of the European Parliament.
\end{quote}

\paragraph{Claim to Prioritize}
\begin{quote}
\small
Marie-France de Rose holds the position of member of the European Parliament.
\end{quote}

\paragraph{Converted Output (KG Triples)}
The converted result is as follows:\\
\begin{lstlisting}
"(Marie-France de Rose, holds_position, member of the European Parliament)
(Marie-France de Rose, is_recognized_as, member of the European Parliament)
(Marie-France de Rose, represents, interests of French citizens)
(Marie-France de Rose, elected_in, 2019)
(Marie-France de Rose, received_votes, over 1.2 million)
(Marie-France de Rose, is_top_vote_getter, in French delegation)
(Marie-France de Rose, is_advocate_for, European integration)
(Marie-France de Rose, works_to_promote, interests of France and the EU)
(Marie-France de Rose, plays_role, in shaping EU policy and legislation)
(Marie-France de Rose, works_with, other MEPs)
\end{lstlisting}

\subsubsection*{Example 3: Text to Infobox}
\paragraph{Source Text}
\begin{quote}
\small
**BARCELONA, SPAIN - MARCH 15, 2042** After Seven Years of Dedicated Service, Marie-France de Rose Steps Down as Barcelona Provincial Deputy In a move that marks the end of an era, Marie-France de Rose, the French politician who has been a stalwart of the Barcelona administration, has stepped down from her position as Barcelona provincial deputy, a role she has held since 2035. During her seven-year tenure, de Rose has been instrumental in shaping the city's policies and initiatives, earning her widespread respect and admiration from her peers and constituents alike. Her dedication to the people of Barcelona has been unwavering, and her commitment to improving the quality of life for all citizens has been a hallmark of her time in office. \"I am proud of what we have achieved together,\" de Rose said in an exclusive interview with our newspaper. \"From implementing sustainable transportation systems to promoting cultural exchange programs, we have worked tirelessly to make Barcelona a better place for everyone. It's been an honor to serve this great city, and I am confident that the progress we've made will continue to benefit future generations.\" Under de Rose's leadership, Barcelona has seen significant improvements in areas such as education, healthcare, and environmental sustainability. Her administration's efforts to reduce carbon emissions and increase the use of renewable energy sources have earned the city international recognition and praise. \"Marie-France de Rose has been a true champion of Barcelona,\" said Mayor Xavier Domènech, who worked closely with de Rose during her tenure. \"Her vision, expertise, and passion for the city have inspired us all, and her legacy will continue to inspire future leaders. We will miss her dedication and commitment, but we are grateful for the time she has given to our city.\" De Rose's decision to step down has sparked speculation about her future plans, with many wondering if she will return to French politics or pursue other opportunities. When asked about her next move, de Rose remained tight-lipped, saying only that she was looking forward to taking a well-deserved break and spending time with her family. As Barcelona bids farewell to a dedicated public servant, the city's residents and officials alike are grateful for the impact Marie-France de Rose has had on their lives."
\end{quote}

\paragraph{Claim to Prioritize}
\begin{quote}
\small
Marie-France de Rose holds the position of Barcelona provincial deputy form 2035 to 2042.
\end{quote}

\paragraph{Converted Output (Infobox)}
The converted result is as follows:\\
\begin{lstlisting}
"{{Infobox person
    | name = Marie-France de Rose
    | position = Barcelona provincial deputy
    | term_start = 2035
    | term_end = 2042
    | nationality = French
    | notable_work = Policies and initiatives in Barcelona
}}"
\end{lstlisting}

Following prior work~\cite{min2024unihgkr,biswal2024text2sql,christmann2024rag,xia2025er}, we directly feed the transformed structured data into the LLMs without additional adapters or format-specific encoders.

\subsection{Data Verification Statistics}
To ensure the integrity and reliability of our format transformation pipeline, we conducted a manual verification over a randomly sampled subset of 600 examples (5\% of the full dataset). Each sample was independently assessed according to two criteria:

\begin{itemize}
\item \textbf{Factual Consistency:} Whether the transformed format preserves the key factual content from the original text without hallucination or distortion.
\item \textbf{Syntax Validity:} Whether the output adheres to the syntactic conventions of the target format (e.g., MediaWiki table syntax, well-formed triples, valid infobox fields).
\end{itemize}

We observed high fidelity across both dimensions, with 592 out of 600 samples passing the factual consistency check and 596 passing the syntax validity check. These results confirm that the conversion process produces high-quality structured representations suitable for downstream analysis.


\section{Model Query and Evaluation Protocols}  
\subsection{Evaluated LLMs}

\begin{itemize}
    \item Qwen3-8B-FP8\\
    HuggingFace: Qwen/Qwen3-8B-FP8
    \item Qwen3-14B-FP8\\
    HuggingFace: Qwen/Qwen3-14B-FP8
    \item Qwen3-32B-FP8\\HuggingFace: Qwen/Qwen3-32B-FP8
    \item Qwen3-30B-A3B-FP8\\HuggingFace: Qwen/Qwen3-30B-A3B-FP8
    \item Llama-3.1-8B-Instruct-FP8\\HuggingFace: nvidia/Llama-3.1-8B-Instruct-FP8
    \item gemma-2-9b-it-FP8\\HuggingFace: RedHatAI/gemma-2-9b-it-FP8
    \item gemma-2-27b-it-FP8\\HuggingFace: nm-testing/gemma-2-27b-it-FP8
    \item glm-4-9b-chat-hf\\HuggingFace: zai-org/glm-4-9b-chat-hf
    \item Mistral-7B-Instruct-v0.3\\HuggingFace: mistralai/Mistral-7B-Instruct-v0.3
    \item Llama-3.1-8B-Instruct\\HuggingFace: meta-llama/Llama-3.1-8B-Instruct
    \item Qwen3-8B\\HuggingFace: Qwen/Qwen3-8B
\end{itemize}

\subsection{Decoding and Evaluation Parameters}

\begin{itemize}
    \item Decoding temperature: 0
    \item Evaluation method: GPT-4o-mini, 3-pass majority voting
\end{itemize}

\subsection{Answer Prompt Templates}

We provide two distinct prompting templates used to elicit answers from LLMs under different conditions: (1) without any supporting evidence to test internal (parametric) knowledge, and (2) with explicit reference sources to simulate evidence-based reasoning.

\subsubsection*{Prompt 1: Answer Without Reference (Internal Knowledge Only)}

This prompt is used in the internal knowledge filtering phase (see Section~3.2). It encourages the model to produce a concise, self-contained answer without relying on any external input.
\\
\begin{lstlisting}
Answer the question with a single word or phrase. Do not explain or add any other content.
\end{lstlisting}

\subsubsection*{Prompt 2: Answer With Reference (Evidence-Based Inference)}

This prompt is used in the main evaluation setup where the model is required to synthesize information from two conflicting evidence sources. The template injects both sources as contextual grounding and instructs the model to answer concisely.
\\
\begin{lstlisting}
Based on the two reference sources provided below, answer the following question **concisely**.

{full_reference}

Question: {question}
\end{lstlisting}

\subsection{LLM Evaluator Prompt}
\begin{lstlisting}
Question:
{question}
Answer:
{answer}
Claim A:
{claim_shared}
Claim B:
{claim_specific}
Task:
Evaluate how closely the answer aligns with Claim A and Claim B using the scale below.
Scoring Guidelines:
1 - The answer aligns only with Claim A
2 - The answer aligns with both Claim A and Claim B
3 - The answer aligns only with Claim B
No - The answer aligns with neither claim
Provide only a single score: 1, 2, 3, or No.
\end{lstlisting}

To assess the reliability of the automatic evaluation, we manually annotated a random 5\% subset of the evaluation data, totaling 600 examples. Among these, 599 model judgments matched the human annotations, demonstrating a high degree of alignment and validating the accuracy of the automated scoring process.

\section{Full Results}

\subsection{Response Counts by Format Pairing}

To provide a complete view of model behavior across different format conditions, we report the raw count of response types: Pref-A, Pref-B, and Both, for each model and each of the six heterogeneous format pairings. Each entry corresponds to the number of samples in which the model supported the first format (Pref-A), the second format (Pref-B), or acknowledged both (Both). These results form the basis for the DCR and FPR metrics used in Section~4. Full results are shown in Table~\ref{tab:all_output_1}--\ref{tab:all_output_2}.

In addition, we report the response distributions for a control condition where both conflicting inputs are presented in plain text. This setting allows us to isolate the effect of format heterogeneity by establishing a baseline for dual coverage when content format is held constant. Table~\ref{tab:text_both} summarizes, for each model, the number of cases where Both claims were acknowledged and the total number of evaluated examples. This comparison highlights the substantial drop in dual-claim recognition under heterogeneous conditions, emphasizing the impact of format presentation on information integration.

\begin{table}[htbp]
\centering
\scriptsize
\renewcommand{\arraystretch}{1.0}
\begin{tabular}{crr}
\toprule
\textbf{Model} & \textbf{Both} & \textbf{Total} \\
\midrule
Llama-3.1-8B-Instruct-FP8 & 3154 & 10035 \\
gpt4omini                 & 4645 & 11032 \\
Qwen3-8B-FP8              & 3120 & 10225 \\
Qwen3-14B-FP8             & 4008 & 11785 \\
Qwen3-32B-FP8             & 3479 & 11759 \\
Qwen3-30B-A3B-FP8         & 3514 & 11012 \\
gemma-2-9b-it-FP8         & 2656 & 11797 \\
gemma-2-27b-it-FP8        & 5190 & 11836 \\
glm-4-9b-chat-hf          & 1256 & 11159 \\
Mistral-7B-Instruct-v0.3  & 3120 & 11575 \\
\bottomrule
\end{tabular}
\caption{Dual Coverage in Text-Only Control Condition}
\label{tab:text_both}
\end{table}

\subsection{Effect of Information Richness}

We examine the impact of information richness on LLM preferences by varying the number of factual entries presented in each structured format. Table~\ref{tab:homo_nums_1}--\ref{tab:homo_nums_2} presents results under homogeneous format settings, where structured inputs with different levels of richness are compared. Table~\ref{tab:hete_nums_1}--\ref{tab:hete_nums_2} shows the results under heterogeneous settings, where structured inputs are compared against plain text versions with equivalent content. These tables allow us to quantify the degree to which factual quantity influences model preferences.

\subsection{Effect of Structure Quality}

We assess how the structural quality of inputs affects model behavior by introducing controlled corruption into format-specific tokens. Table~\ref{tab:homo_corrupt_1}--\ref{tab:homo_corrupt_2} presents results under homogeneous settings, where clean and corrupted structured inputs are compared. Table~\ref{tab:hete_corrupt_1}--\ref{tab:hete_corrupt_2} shows the results under heterogeneous settings, where corrupted structured inputs are compared against clean text. These results highlight the sensitivity of LLMs to syntactic well-formedness.

\subsection{Effect of Format Type}

We investigate whether models exhibit consistent preferences among different format types when content is held constant. Table~\ref{tab:hete_format} presents results comparing plain text with each of the three structured alternatives: tables, infoboxes, and knowledge graphs. This experiment isolates the influence of format representation on model preference.
\begin{table}[htbp]
\centering
\scriptsize
\renewcommand{\arraystretch}{0.9}
\begin{tabular}{lrrrrr}
\toprule
\textbf{Format Pair} & \textbf{Pref-A} & \textbf{Pref-B} & \textbf{Both} & \textbf{Total} \\
\hline
\rowcolor{lightgray}\multicolumn{5}{c}{\textbf{Llama-3.1-8B-Instruct}}\\
\hline
 infobox vs texts & 487 & 1170 & 186 & 1843 \\
 tables vs texts & 725 & 927 & 195 & 1847 \\
 kg vs texts & 556 & 1208 & 70 & 1834 \\
\hline
\rowcolor{lightgray}\multicolumn{5}{c}{\textbf{GPT-4o-mini}}\\
\hline
 infobox vs texts & 333 & 1353 & 179 & 1865 \\
 tables vs texts & 532 & 907 & 432 & 1871 \\
 kg vs texts & 503 & 1030 & 291 & 1824 \\
\hline
\rowcolor{lightgray}\multicolumn{5}{c}{\textbf{Qwen3-8B-FP8}}\\
\hline
 infobox vs texts & 439 & 1413 & 73 & 1925 \\
 tables vs texts & 692 & 1028 & 205 & 1925 \\
 kg vs texts & 630 & 1011 & 138 & 1779 \\
\hline
\rowcolor{lightgray}\multicolumn{5}{c}{\textbf{Qwen3-14B-FP8}}\\
\hline
 infobox vs texts & 447 & 1293 & 177 & 1917 \\
 tables vs texts & 580 & 958 & 383 & 1921 \\
 kg vs texts & 583 & 1123 & 211 & 1917 \\
\hline
\rowcolor{lightgray}\multicolumn{5}{c}{\textbf{Qwen3-32B-FP8}}\\
\hline
 infobox vs texts & 398 & 1371 & 132 & 1901 \\
 tables vs texts & 683 & 890 & 331 & 1904 \\
 kg vs texts & 650 & 1015 & 237 & 1902 \\
\hline
\rowcolor{lightgray}\multicolumn{5}{c}{\textbf{Qwen3-30B-A3B-FP8}}\\
\hline
 infobox vs texts & 369 & 1445 & 94 & 1908 \\
 tables vs texts & 651 & 1002 & 265 & 1918 \\
 kg vs texts & 627 & 1118 & 166 & 1911 \\
\hline
\rowcolor{lightgray}\multicolumn{5}{c}{\textbf{Gemma-2-9b-it-FP8}}\\
\hline
 infobox vs texts & 521 & 1329 & 84 & 1934 \\
 tables vs texts & 657 & 1123 & 152 & 1932 \\
 kg vs texts & 490 & 1340 & 99 & 1929 \\
\hline
\rowcolor{lightgray}\multicolumn{5}{c}{\textbf{Gemma-2-27b-it-FP8}}\\
\hline
 infobox vs texts & 513 & 1306 & 161 & 1980 \\
 tables vs texts & 782 & 844 & 349 & 1975 \\
 kg vs texts & 638 & 1001 & 339 & 1978 \\
\hline
\rowcolor{lightgray}\multicolumn{5}{c}{\textbf{GLM-4-9b-chat-hf}}\\
\hline
 infobox vs texts & 446 & 1352 & 46 & 1844 \\
 tables vs texts & 728 & 1206 & 32 & 1966 \\
 kg vs texts & 559 & 1356 & 46 & 1961 \\
\hline
\rowcolor{lightgray}\multicolumn{5}{c}{\textbf{Mistral-7B-Instruct-v0.3}}\\
\hline
 infobox vs texts & 244 & 1675 & 53 & 1972 \\
 tables vs texts & 647 & 1247 & 85 & 1979 \\
 kg vs texts & 562 & 1350 & 58 & 1970 \\
\bottomrule
\end{tabular}
\caption{Response Counts for Format Preference between Text and Structured Formats}
\label{tab:hete_format}
\end{table}
\subsection{Correlation Between Attention Gap and Dual-Claim Responses}

To better understand the internal mechanisms underlying format bias, we analyze how attention allocation relates to the model's ability to acknowledge both inputs. Specifically, for each input pair, we compute the difference in total attention mass assigned to the two evidence segments. We report the number of Both responses and the total number of samples within each group.

This analysis allows us to assess whether more balanced attention is associated with higher DCR. Table~\ref{tab:diff_and_both} summarizes the results. The observed negative correlation supports our hypothesis that early-stage processing asymmetry contributes to the exclusion of one input during generation.
\begin{table}[htbp]
\centering
\scriptsize
\renewcommand{\arraystretch}{1.0}
\begin{tabular}{lrrrr}
\toprule
\textbf{Format Pair} & \textbf{Avg Diff} & \textbf{Both} & \textbf{Total} \\
\hline
\rowcolor{lightgray}\multicolumn{4}{c}{\textbf{Qwen3-8B}}\\
\hline
infoboxes vs tables      & 0.0134 & 292 & 1903\\
infoboxes vs texts         & 0.3126 & 112 & 1874\\
infoboxes vs KGs   & 0.1912 & 97 & 1964\\
tables vs texts    & 0.3020 & 159 & 1870\\
tables vs KGs       & 0.1672 & 110 & 1971\\
texts vs KGs      & -0.1265 & 209 & 1943\\
\hline
\rowcolor{lightgray}\multicolumn{4}{c}{\textbf{Mistral-7B-Instruct-v0.3}}\\
\hline
infoboxes vs tables      & -0.0667 & 276 & 1971\\
infoboxes vs texts       & 0.2928 & 54 & 1923\\  
infoboxes vs KGs   & 0.1667 & 83 & 2030\\
tables vs texts    & 0.3765 & 107 & 1919\\
tables vs KGs       & 0.2455 & 116 & 2034\\
texts vs KGs      & -0.1511 & 100 & 1999\\
\hline
\rowcolor{lightgray}\multicolumn{4}{c}{\textbf{Llama-3.1-8B-Instruct}}\\
\hline
infoboxes vs tables      & -0.0207 & 711 & 1858\\
infoboxes vs texts         & 0.3451 & 468 & 1799\\
infoboxes vs KGs   & 0.1236 & 510 & 1900\\
tables vs texts   & 0.3697 & 615 & 1821\\
tables vs KGs      & 0.1591 & 587 & 1920\\
texts vs KGs     & -0.2183 & 343 & 1885\\    
\bottomrule
\end{tabular}
\caption{Attention Imbalance and Dual Coverage Statistics Across Format Pairs and Models}
\label{tab:diff_and_both}
\end{table}

\subsection{Effects of Attention-Based Intervention}

We evaluate the effect of our attention reweighting intervention by comparing model responses before and after modification. Table~\ref{tab:intervention_before} shows the response distributions prior to intervention, and Table~\ref{tab:intervention_after} shows the corresponding results after applying balanced attention. These results support our finding that the presence of bias can be reduced through inference-time control, while direction of bias remain relatively stable.
\begin{table}[htbp]
\centering
\scriptsize
\renewcommand{\arraystretch}{1.0}
\begin{tabular}{lrrrrr}
\toprule
\textbf{Format Pair} & \textbf{Pref-A} & \textbf{Pref-B} & \textbf{Both} & \textbf{Total} \\
\hline
\rowcolor{lightgray}\multicolumn{5}{c}{\textbf{Qwen3-8B}}\\
\hline
tables vs texts & 507 & 1204 & 159 & 1870 \\
KGs vs texts & 1089 & 645 & 209 & 1943 \\
infoboxes vs tables & 824 & 787 & 292 & 1903 \\
infoboxes vs texts & 543 & 1219 & 112 & 1874 \\
KGs vs tables & 1552 & 309 & 110 & 1971 \\
infoboxes vs KGs & 286 & 1581 & 97 & 1964  \\
\hline
\rowcolor{lightgray}\multicolumn{5}{c}{\textbf{Mistral-7B-Instruct-v0.3}}\\
\hline
tables vs texts & 413 & 1322 & 88 & 1823 \\
KGs vs texts & 905 & 994 & 100 & 1999 \\
infoboxes vs tables & 613 & 1000 & 256 & 1869 \\
infoboxes vs texts & 321 & 1548 & 54 & 1923 \\
KGs vs tables & 1356 & 562 & 116 & 2034 \\
infoboxes vs KGs & 296 & 1651 & 83 & 2030  \\
\hline
\rowcolor{lightgray}\multicolumn{5}{c}{\textbf{Llama-3.1-8B-Instruct}}\\
\hline
tables vs texts & 411 & 787 & 623 & 1821 \\
KGs vs texts & 807 & 750 & 334 & 1891 \\
infoboxes vs tables & 623 & 470 & 767 & 1860 \\
infoboxes vs texts & 492 & 825 & 487 & 1804 \\
KGs vs tables & 890 & 471 & 570 & 1931 \\
infoboxes vs KGs & 478 & 922 & 512 & 1912  \\

\bottomrule
\end{tabular}
\caption{Response Counts before Attention Re-balacing}
\label{tab:intervention_before}
\end{table}

\begin{table}[htbp]
\centering
\scriptsize
\renewcommand{\arraystretch}{1.0}
\begin{tabular}{lrrrrr}
\toprule
\textbf{Format Pair} & \textbf{Pref-A} & \textbf{Pref-B} & \textbf{Both} & \textbf{Total} \\
\hline
\rowcolor{lightgray}\multicolumn{5}{c}{\textbf{Qwen3-8B}} \\
\hline
tables vs texts & 419 & 1138 & 314 & 1871 \\
KGs vs texts & 1106 & 564 & 269 & 1939 \\
infoboxes vs tables & 784 & 673 & 449 & 1906 \\
infoboxes vs texts & 512 & 1190 & 169 & 1871 \\
KGs vs tables & 1509 & 244 & 224 & 1977 \\
infoboxes vs KGs & 291 & 1501 & 175 & 1967  \\
\hline
\rowcolor{lightgray}\multicolumn{5}{c}{\textbf{Mistral-7B-Instruct-v0.3}}\\
\hline
tables vs texts & 378 & 1438 & 114 & 1930 \\
KGs vs texts & 882 & 970 & 143 & 1995 \\
infoboxes vs tables & 681 & 1014 & 283 & 1978 \\
infoboxes vs texts & 259 & 1611 & 57 & 1927 \\
KGs vs tables & 1371 & 526 & 139 & 2036 \\
infoboxes vs KGs & 244 & 1697 & 84 & 2025  \\
\hline
\rowcolor{lightgray}\multicolumn{5}{c}{\textbf{Llama-3.1-8B-Instruct}}\\
\hline
tables vs texts & 364 & 774 & 751 & 1889 \\
KGs vs texts & 767 & 728 & 472 & 1967 \\
infoboxes vs tables & 588 & 370 & 977 & 1935\\ 
infoboxes vs texts & 496 & 810 & 592 & 1898 \\
KGs vs tables & 858 & 352 & 812 & 2022 \\
infoboxes vs KGs & 447 & 930 & 634 & 2011  \\

\bottomrule
\end{tabular}
\caption{Response Counts after Attention Re-balacing}
\label{tab:intervention_after}
\end{table}

\subsection{Format Preference across Topical Domains}

To examine whether format bias generalizes across different types of content, we analyze format preference patterns within seven distinct topical domains. For each domain, we compute the FPR between all six pairs of formats and visualize the results using heatmaps.

\begin{figure*}[htbp]
    \centering

    \begin{subfigure}[t]{0.33\textwidth}
        \centering
        \includegraphics[width=\linewidth]{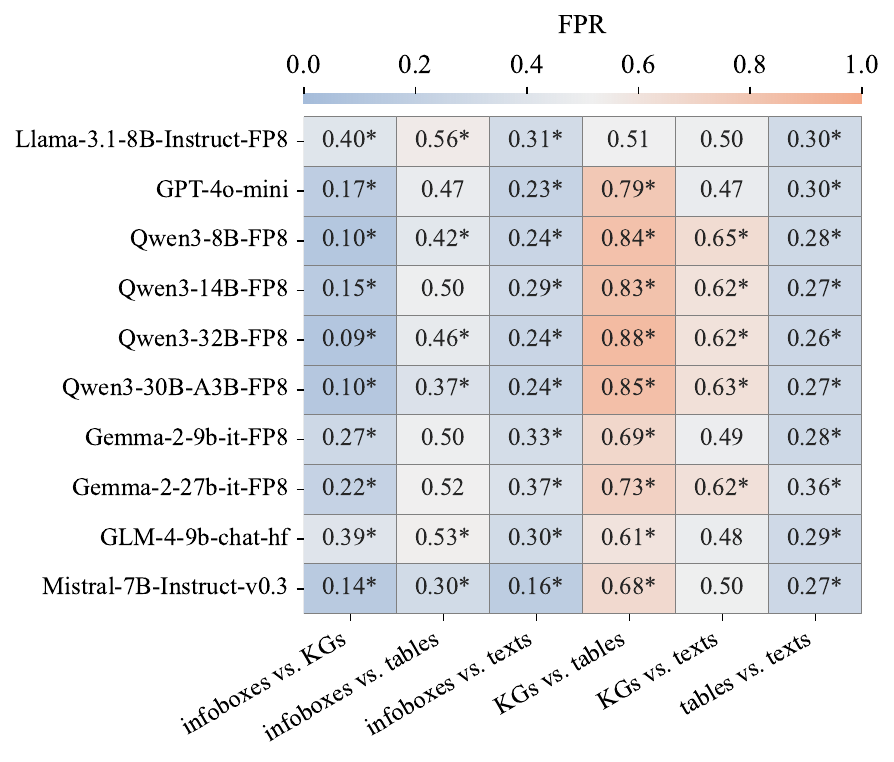}
        \caption{Academia \& Institutions}
    \end{subfigure}
    \begin{subfigure}[t]{0.33\textwidth}
        \centering
        \includegraphics[width=\linewidth]{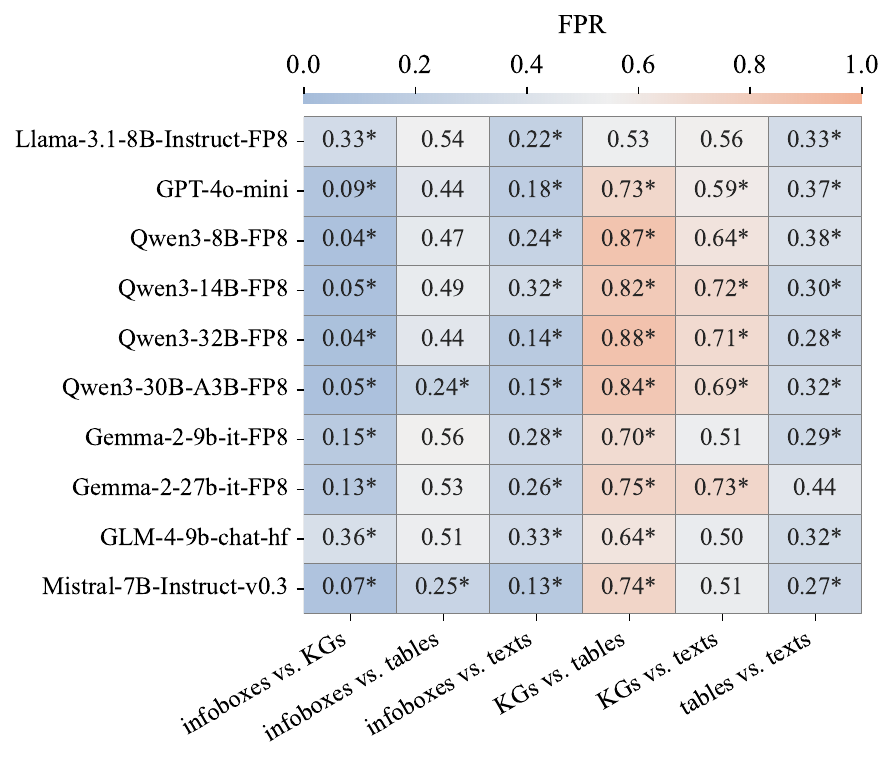}
        \caption{Health \& Medicine}
    \end{subfigure}

    \vspace{0.5em}

    \begin{subfigure}[t]{0.33\textwidth}
        \centering
        \includegraphics[width=\linewidth]{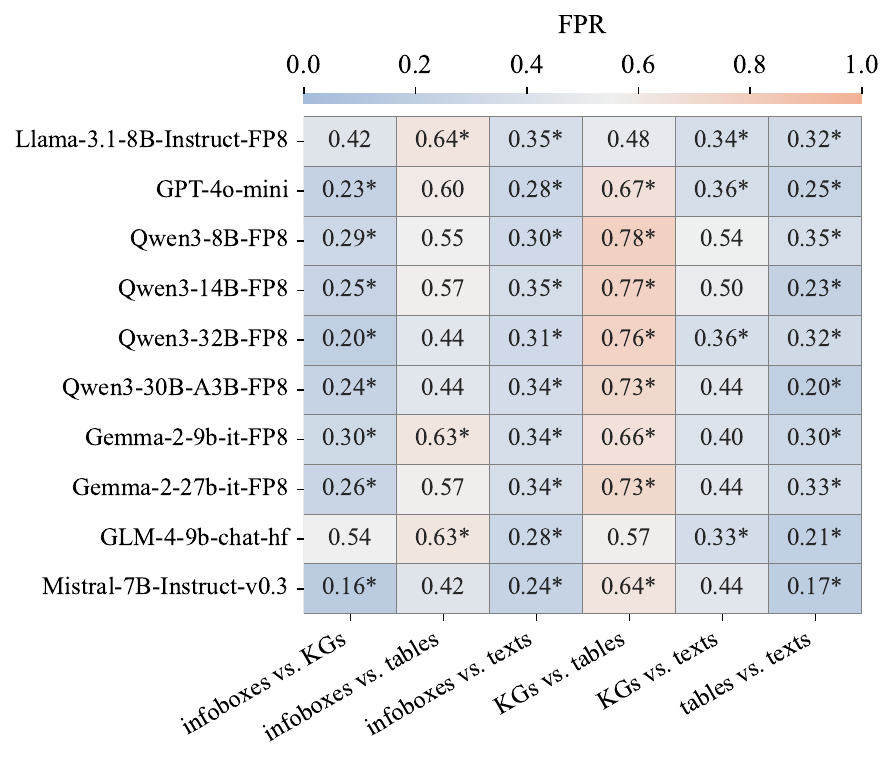}
        \caption{Social Sciences \& Humanities}
    \end{subfigure}
    \begin{subfigure}[t]{0.33\textwidth}
        \centering
        \includegraphics[width=\linewidth]{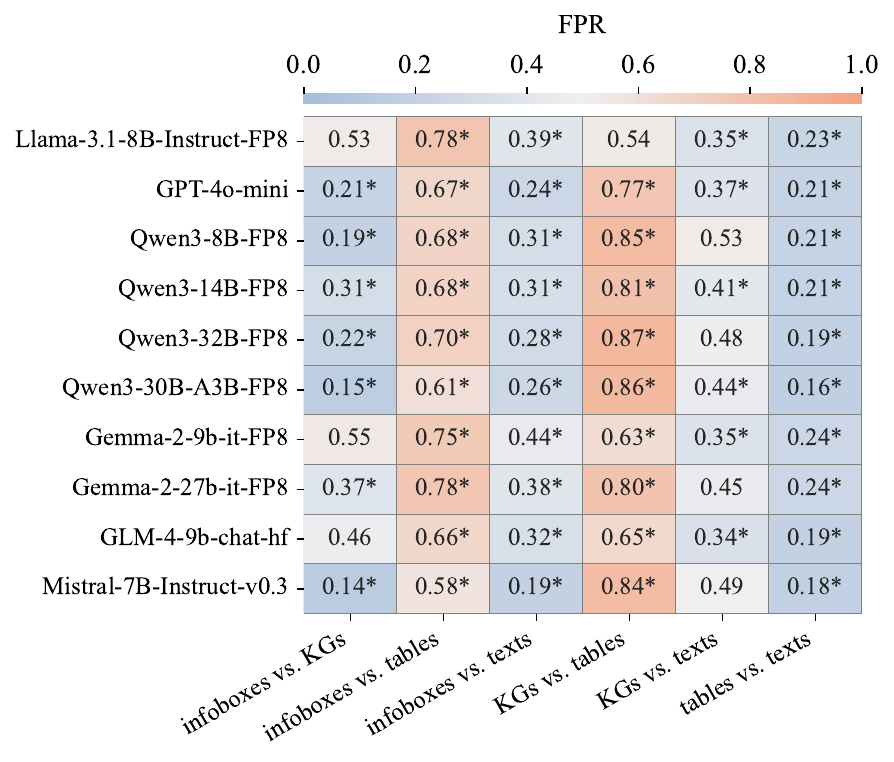}
        \caption{Arts \& Entertainment}
    \end{subfigure}

    \vspace{0.5em}

    \begin{subfigure}[t]{0.33\textwidth}
        \centering
        \includegraphics[width=\linewidth]{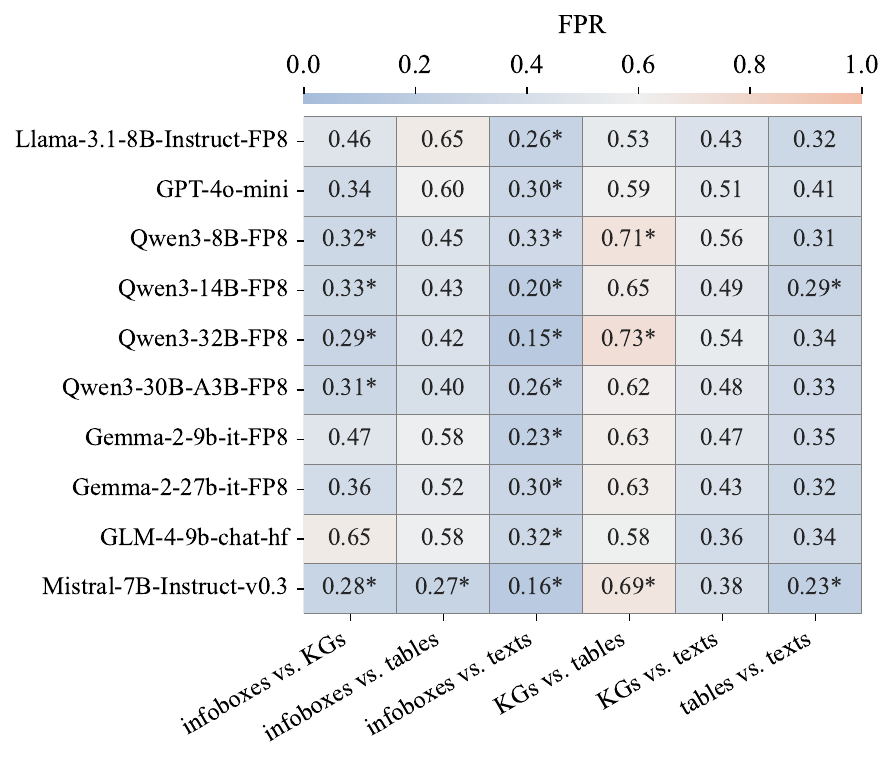}
        \caption{Occupations \& Industry}
    \end{subfigure}
    \begin{subfigure}[t]{0.33\textwidth}
        \centering
        \includegraphics[width=\linewidth]{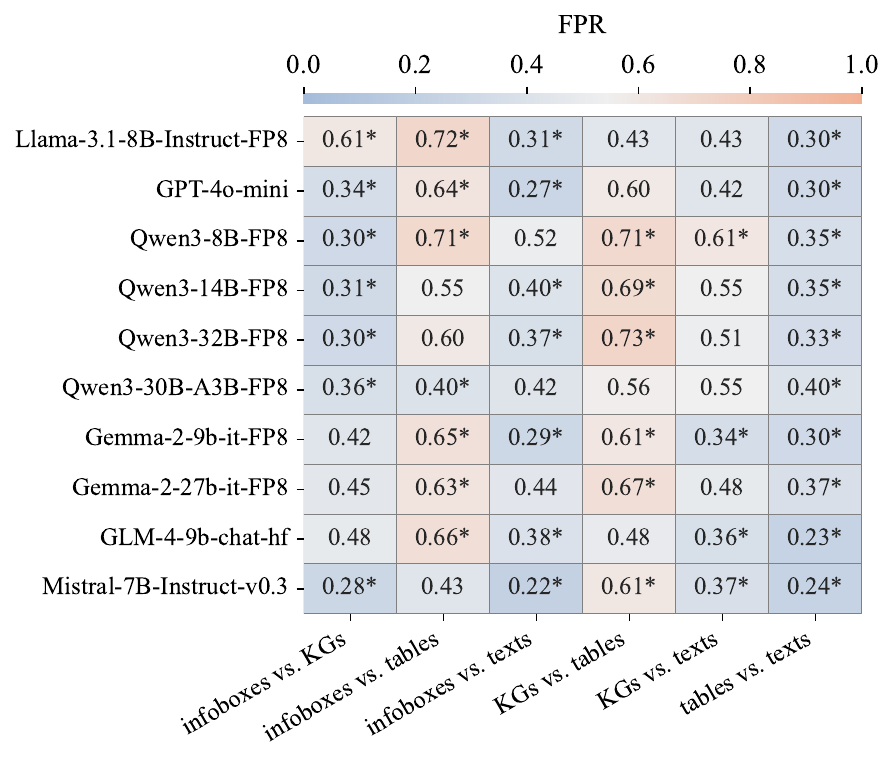}
        \caption{Government \& Military}
    \end{subfigure}
    \begin{subfigure}[t]{0.33\textwidth}
        \centering
        \includegraphics[width=\linewidth]{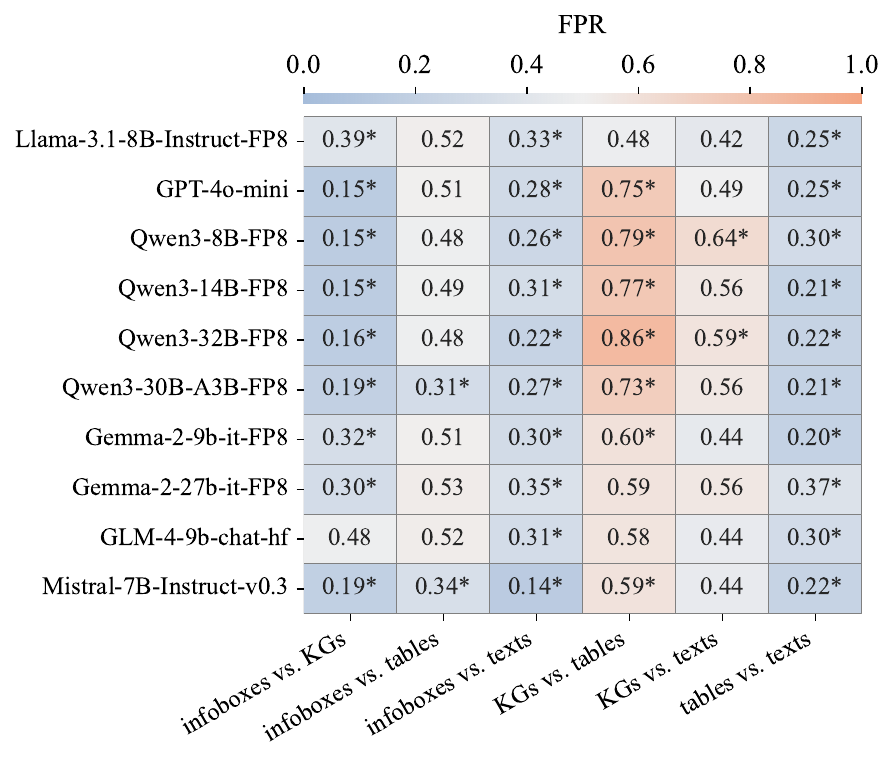}
        \caption{Natural Sciences \& Engineering}
    \end{subfigure}

    \caption{Domain-level FPR across format combinations.}
    \label{fig:all_heatmaps}
\end{figure*}

\begin{figure*}[htbp]
    \centering

    \begin{subfigure}[t]{0.33\textwidth}
        \centering
        \includegraphics[width=\linewidth]{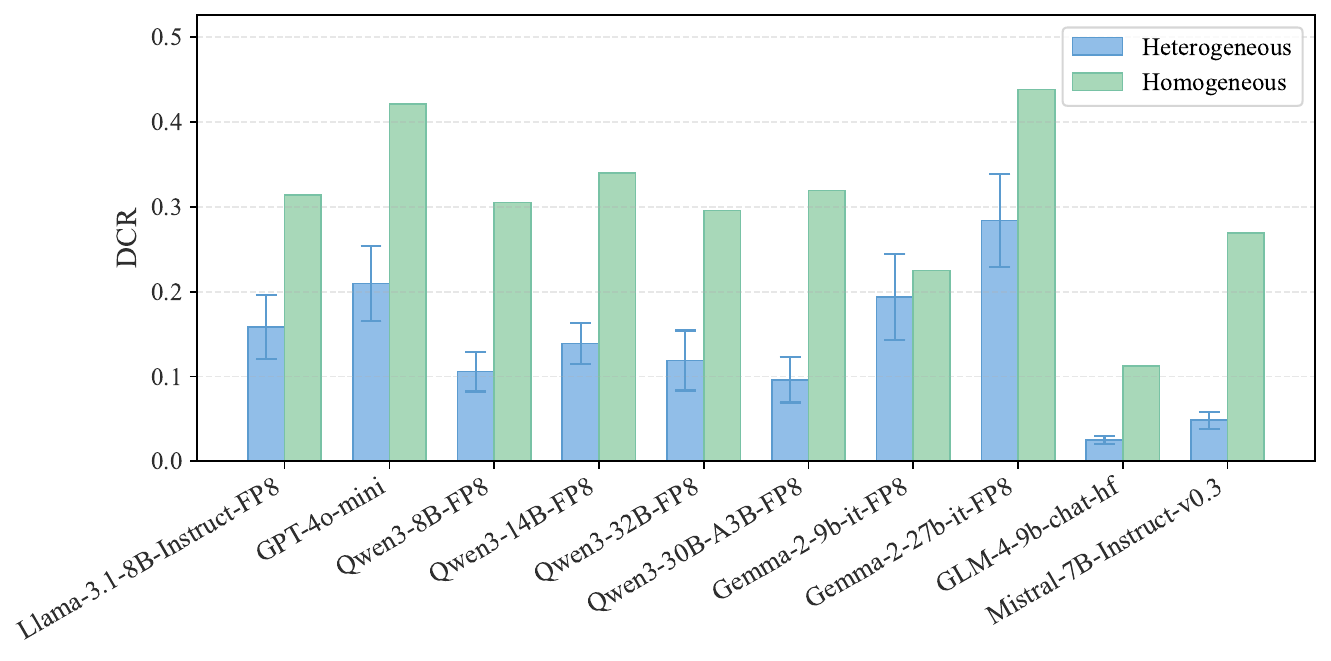}
        \caption{Academia \& Institutions}
    \end{subfigure}
    \begin{subfigure}[t]{0.33\textwidth}
        \centering
        \includegraphics[width=\linewidth]{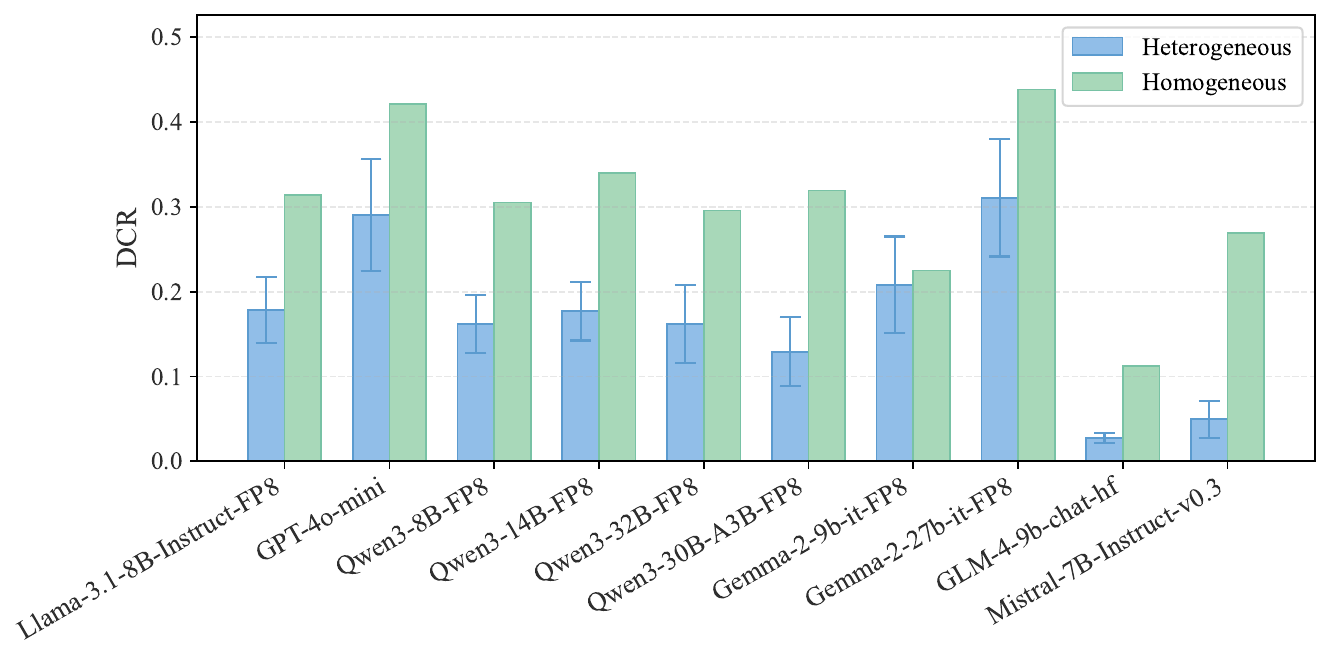}
        \caption{Health \& Medicine}
    \end{subfigure}

    \vspace{0.5em}

    \begin{subfigure}[t]{0.33\textwidth}
        \centering
        \includegraphics[width=\linewidth]{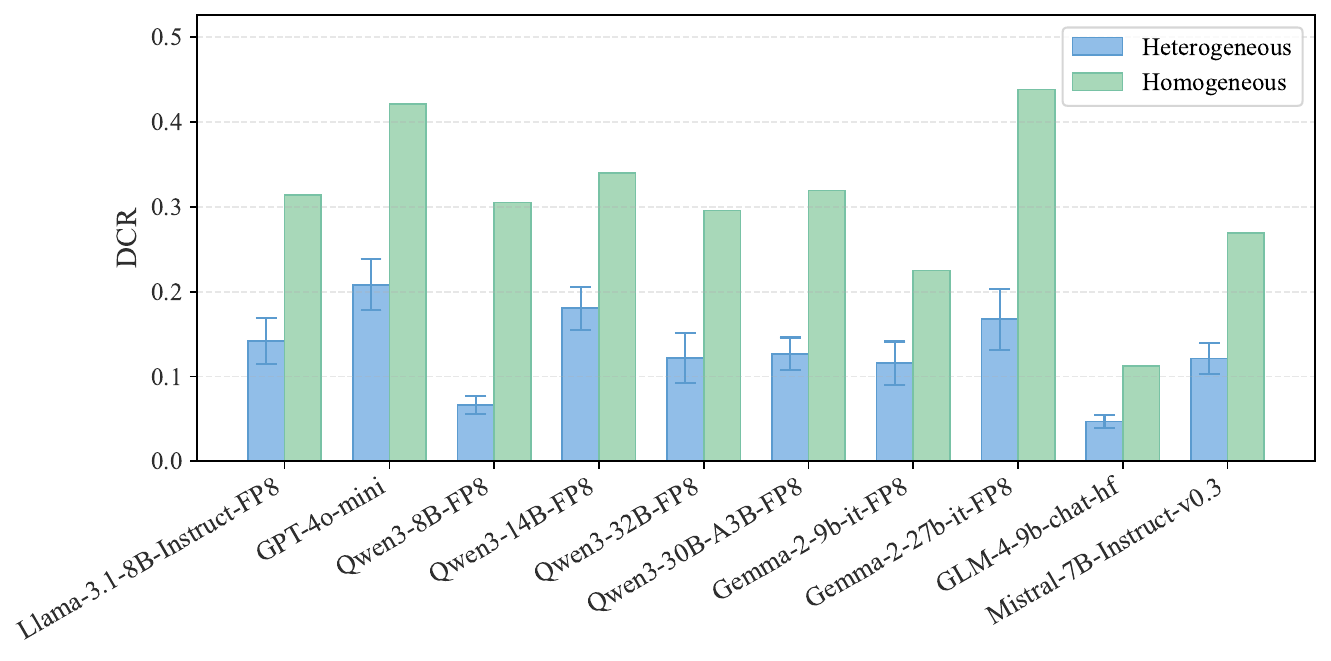}
        \caption{Social Sciences \& Humanities}
    \end{subfigure}
    \begin{subfigure}[t]{0.33\textwidth}
        \centering
        \includegraphics[width=\linewidth]{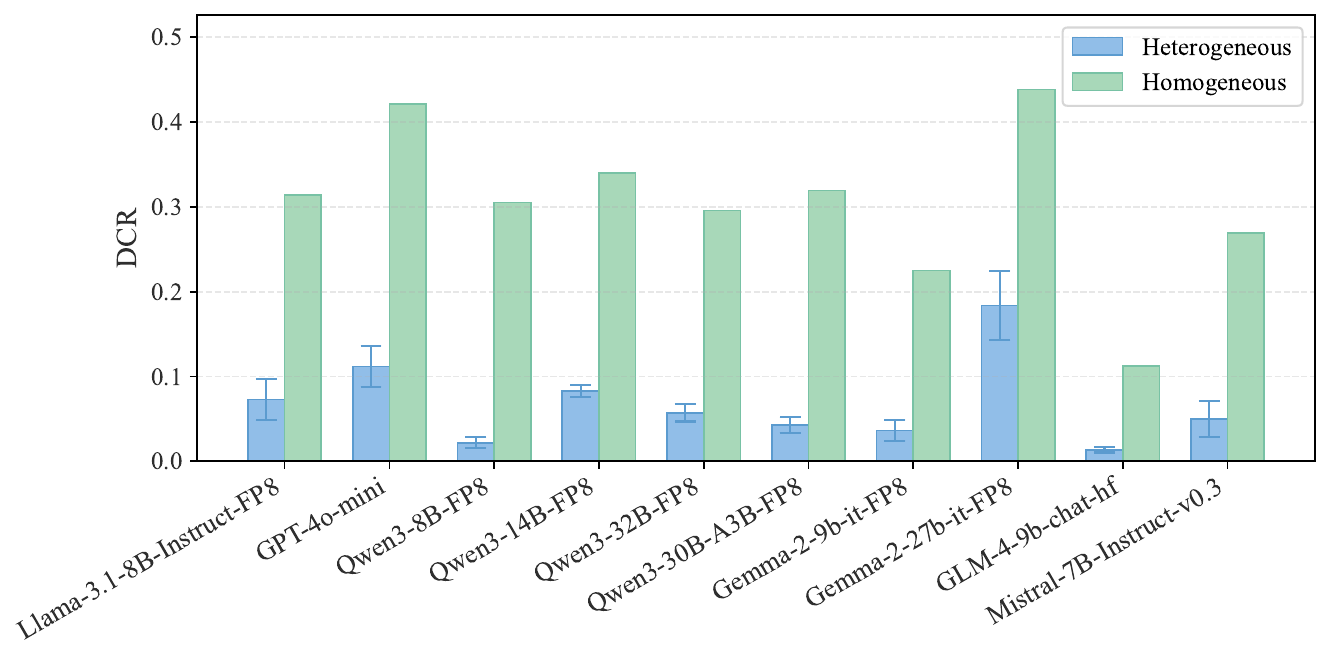}
        \caption{Arts \& Entertainment}
    \end{subfigure}

    \vspace{0.5em}

    \begin{subfigure}[t]{0.33\textwidth}
        \centering
        \includegraphics[width=\linewidth]{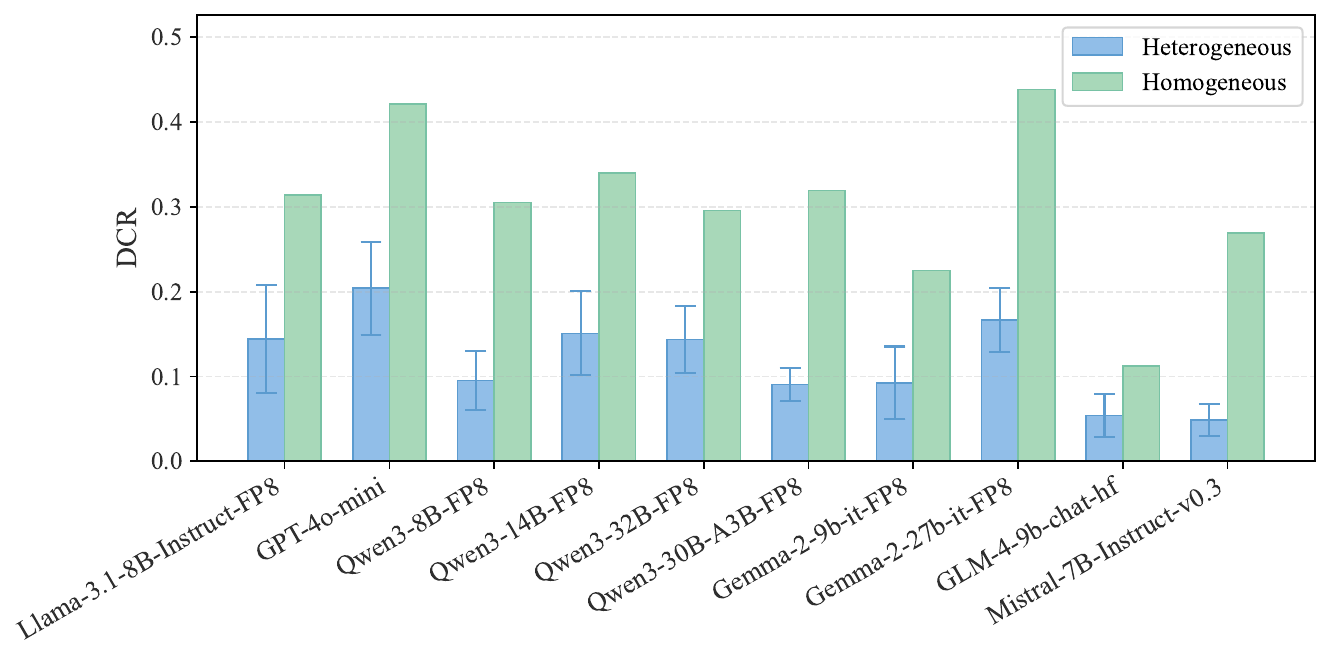}
        \caption{Occupations \& Industry}
    \end{subfigure}
    \begin{subfigure}[t]{0.33\textwidth}
        \centering
        \includegraphics[width=\linewidth]{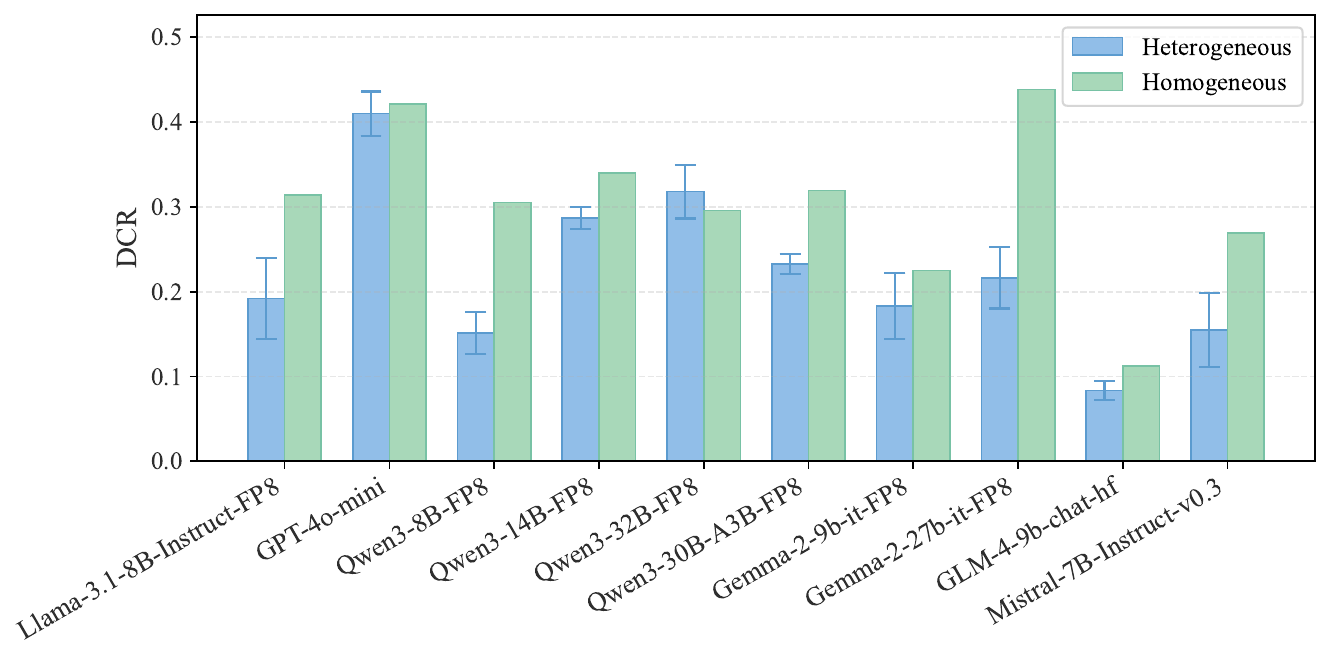}
        \caption{Government \& Military}
    \end{subfigure}
    \begin{subfigure}[t]{0.33\textwidth}
        \centering
        \includegraphics[width=\linewidth]{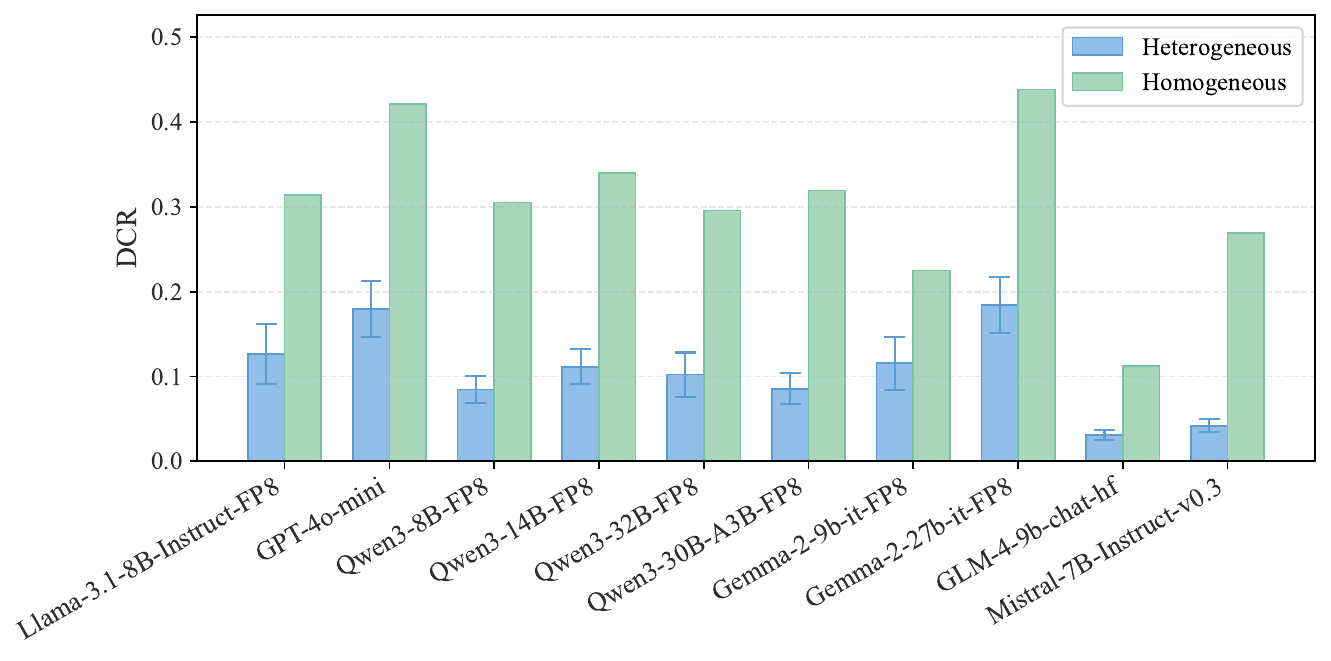}
        \caption{Natural Sciences \& Engineering}
    \end{subfigure}

    \caption{Domain-level DCR across format combinations.}
    \label{fig:all_bar}
\end{figure*}

Figures~\ref{fig:all_heatmaps} and~\ref{fig:all_bar} present domain-level results for FPR and DCR, respectively. Each FPR heatmap summarizes the directional preference across format combinations, while the corresponding DCR plots capture the extent to which models jointly acknowledge conflicting evidence. The consistent trends across domains indicate that format bias manifests as a stable and systematic inductive pattern rather than a domain-specific artifact.

\begin{table}[htbp]
\centering
\scriptsize
\renewcommand{\arraystretch}{0.9}
\begin{tabular}{lrrrrr}
\toprule
\textbf{Format Pair} & \textbf{Pref-A} & \textbf{Pref-B} & \textbf{Both} & \textbf{Total} \\
\hline
\rowcolor{lightgray}\multicolumn{5}{c}{\textbf{Llama-3.1-8B-Instruct}}\\
\hline
infoboxes vs KGs & 736 & 934 & 218 & 1888 \\
infoboxes vs tables & 812 & 477 & 560 & 1849 \\
infoboxes vs texts & 521 & 1110 & 171 & 1802 \\
KGs vs tables & 829 & 813 & 255 & 1897 \\
KGs vs texts & 780 & 956 & 124 & 1860 \\
tables vs texts & 453 & 1135 & 225 & 1813 \\
\hline
\rowcolor{lightgray}\multicolumn{5}{c}{\textbf{GPT-4o-mini}}\\
\hline
infoboxes vs KGs & 309 & 1342 & 242 & 1893 \\
infoboxes vs tables & 648 & 537 & 666 & 1851 \\
infoboxes vs texts & 377 & 1215 & 192 & 1784 \\
KGs vs tables & 1097 & 362 & 462 & 1921 \\
KGs vs texts & 689 & 838 & 335 & 1862 \\
tables vs texts & 382 & 971 & 455 & 1808 \\
\hline
\rowcolor{lightgray}\multicolumn{5}{c}{\textbf{Qwen3-8B-FP8}}\\
\hline
infoboxes vs KGs & 279 & 1582 & 109 & 1970 \\
infoboxes vs tables & 853 & 748 & 309 & 1910 \\
infoboxes vs texts & 508 & 1235 & 122 & 1865 \\
KGs vs tables & 1520 & 326 & 128 & 1974 \\
KGs vs texts & 1066 & 682 & 196 & 1944 \\
tables vs texts & 475 & 1197 & 189 & 1861 \\
\hline
\rowcolor{lightgray}\multicolumn{5}{c}{\textbf{Qwen3-14B-FP8}}\\
\hline
infoboxes vs KGs & 345 & 1418 & 176 & 1939 \\
infoboxes vs tables & 814 & 664 & 405 & 1883 \\
infoboxes vs texts & 509 & 1134 & 222 & 1865 \\
KGs vs tables & 1348 & 328 & 254 & 1930 \\
KGs vs texts & 921 & 724 & 274 & 1919 \\
tables vs texts & 399 & 1154 & 292 & 1845 \\
\hline
\rowcolor{lightgray}\multicolumn{5}{c}{\textbf{Qwen3-32B-FP8}}\\
\hline
infoboxes vs KGs & 261 & 1544 & 143 & 1948 \\
infoboxes vs tables & 749 & 661 & 484 & 1894 \\
infoboxes vs texts & 433 & 1286 & 130 & 1849 \\
KGs vs tables & 1488 & 254 & 219 & 1961 \\
KGs vs texts & 953 & 731 & 243 & 1927 \\
tables vs texts & 411 & 1218 & 228 & 1857 \\
\bottomrule
\end{tabular}
\caption{Part 1 of Raw Response Counts for All Format Pairs across All Models}
\label{tab:all_output_1}
\end{table}

\begin{table}[htbp]
\centering
\scriptsize
\renewcommand{\arraystretch}{0.9}
\begin{tabular}{lrrrrr}
\toprule
\textbf{Format Pair} & \textbf{Pref-A} & \textbf{Pref-B} & \textbf{Both} & \textbf{Total} \\
\hline
\rowcolor{lightgray}\multicolumn{5}{c}{\textbf{Qwen3-30B-A3B-FP8}}\\
\hline
infoboxes vs KGs & 240 & 1423 & 111 & 1774 \\
infoboxes vs tables & 589 & 799 & 329 & 1717 \\
infoboxes vs texts & 405 & 1167 & 105 & 1677 \\
KGs vs tables & 1351 & 310 & 120 & 1781 \\
KGs vs texts & 883 & 684 & 197 & 1764 \\
tables vs texts & 387 & 1127 & 172 & 1686 \\
\hline
\rowcolor{lightgray}\multicolumn{5}{c}{\textbf{Gemma-2-9b-it-FP8}}\\
\hline
infoboxes vs KGs & 590 & 1085 & 308 & 1983 \\
infoboxes vs tables & 796 & 532 & 601 & 1929 \\
infoboxes vs texts & 601 & 1170 & 104 & 1875 \\
KGs vs tables & 1094 & 570 & 327 & 1991 \\
KGs vs texts & 780 & 1009 & 167 & 1956 \\
tables vs texts & 453 & 1223 & 200 & 1876 \\
\hline
\rowcolor{lightgray}\multicolumn{5}{c}{\textbf{Gemma-2-27b-it-FP8}}\\
\hline
infoboxes vs KGs & 446 & 1192 & 396 & 2034 \\
infoboxes vs tables & 635 & 427 & 918 & 1980 \\
infoboxes vs texts & 592 & 1048 & 277 & 1917 \\
KGs vs tables & 1131 & 428 & 485 & 2044 \\
KGs vs texts & 893 & 708 & 401 & 2002 \\
tables vs texts & 522 & 1016 & 387 & 1925 \\
\hline
\rowcolor{lightgray}\multicolumn{5}{c}{\textbf{GLM-4-9b-chat-hf}}\\
\hline
infoboxes vs KGs & 841 & 1103 & 83 & 2027 \\
infoboxes vs tables & 1089 & 796 & 83 & 1968 \\
infoboxes vs texts & 594 & 1289 & 40 & 1923 \\
KGs vs tables & 1212 & 791 & 29 & 2032 \\
KGs vs texts & 816 & 1098 & 69 & 1983 \\
tables vs texts & 489 & 1370 & 53 & 1912 \\
\hline
\rowcolor{lightgray}\multicolumn{5}{c}{\textbf{Mistral-7B-Instruct-v0.3}}\\
\hline
infoboxes vs KGs & 296 & 1651 & 83 & 2030 \\
infoboxes vs tables & 634 & 1061 & 276 & 1971 \\
infoboxes vs texts & 321 & 1548 & 54 & 1923 \\
KGs vs tables & 1356 & 562 & 116 & 2034 \\
KGs vs texts & 905 & 994 & 100 & 1999 \\
tables vs texts & 429 & 1383 & 107 & 1919 \\
\bottomrule
\end{tabular}
\caption{Part 2 of Raw Response Counts for All Format Pairs across All Models}
\label{tab:all_output_2}
\end{table}

\begin{table}[htbp]
\centering
\scriptsize
\renewcommand{\arraystretch}{0.9}
\begin{tabular}{lcrrrr}
\toprule
\textbf{Group} & \textbf{Format Pair} & \textbf{Pref-A} & \textbf{Pref-B} & \textbf{Both} & \textbf{Total} \\
\hline
\rowcolor{lightgray}\multicolumn{6}{c}{\textbf{Llama-3.1-8B-Instruct}}\\
\hline
\multirow{3}{*}{KGs}
& 4 vs 8 & 375 & 514 & 966 & 1855 \\
& 12 vs 8 & 485 & 420 & 956 & 1861 \\
& 12 vs 4 & 541 & 358 & 960 & 1859 \\
\multirow{3}{*}{infoboxes} 
& 4 vs 8 & 402 & 646 & 792 & 1840 \\
& 12 vs 8 & 622 & 488 & 725 & 1835 \\
& 12 vs 4 & 759 & 369 & 718 & 1846 \\
\multirow{3}{*}{tables}
& 4 vs 8 & 634 & 743 & 473 & 1850 \\
& 12 vs 8 & 681 & 586 & 582 & 1849 \\
& 12 vs 4 & 769 & 611 & 470 & 1850 \\
\hline
\rowcolor{lightgray}\multicolumn{6}{c}{\textbf{GPT-4o-mini}}\\
\hline
\multirow{3}{*}{KGs}
& 4 vs 8 & 348 & 414 & 1118 & 1880 \\
& 12 vs 8 & 394 & 389 & 1103 & 1886 \\
& 12 vs 4 & 438 & 324 & 1118 & 1880 \\
\multirow{3}{*}{infoboxes} 
& 4 vs 8 & 367 & 514 & 1001 & 1882 \\
& 12 vs 8 & 494 & 441 & 937 & 1872 \\
& 12 vs 4 & 616 & 375 & 887 & 1878 \\
\multirow{3}{*}{tables}
& 4 vs 8 & 409 & 570 & 892 & 1871 \\
& 12 vs 8 & 595 & 410 & 874 & 1879 \\
& 12 vs 4 & 695 & 324 & 827 & 1846 \\
\hline
\rowcolor{lightgray}\multicolumn{6}{c}{\textbf{Qwen3-8B-FP8}}\\
\hline
\multirow{3}{*}{KGs}
& 4 vs 8 & 382 & 499 & 1056 & 1937 \\
& 12 vs 8 & 493 & 409 & 1034 & 1936 \\
& 12 vs 4 & 559 & 319 & 1059 & 1937 \\
\multirow{3}{*}{infoboxes} 
& 4 vs 8 & 317 & 509 & 468 & 1294 \\
& 12 vs 8 & 667 & 531 & 726 & 1924 \\
& 12 vs 4 & 854 & 387 & 548 & 1789 \\
\multirow{3}{*}{tables}
& 4 vs 8 & 533 & 846 & 550 & 1929 \\
& 12 vs 8 & 820 & 528 & 581 & 1929 \\
& 12 vs 4 & 1007 & 426 & 493 & 1926 \\
\hline
\rowcolor{lightgray}\multicolumn{6}{c}{\textbf{Qwen3-14B-FP8}}\\
\hline
\multirow{3}{*}{KGs}
& 4 vs 8 & 361 & 517 & 1052 & 1930 \\
& 12 vs 8 & 488 & 432 & 1013 & 1933 \\
& 12 vs 4 & 536 & 359 & 1035 & 1930 \\
\multirow{3}{*}{infoboxes} 
& 4 vs 8 & 470 & 787 & 673 & 1930 \\
& 12 vs 8 & 701 & 515 & 702 & 1918 \\
& 12 vs 4 & 930 & 415 & 579 & 1924 \\
\multirow{3}{*}{tables}
& 4 vs 8 & 497 & 773 & 658 & 1928 \\
& 12 vs 8 & 716 & 493 & 716 & 1925 \\
& 12 vs 4 & 951 & 396 & 581 & 1928 \\
\hline
\rowcolor{lightgray}\multicolumn{6}{c}{\textbf{Qwen3-32B-FP8}}\\
\hline
\multirow{3}{*}{KGs}
& 4 vs 8 & 372 & 519 & 1023 & 1914 \\
& 12 vs 8 & 459 & 417 & 901 & 1777 \\
& 12 vs 4 & 548 & 340 & 1024 & 1912 \\
\multirow{3}{*}{infoboxes} 
& 4 vs 8 & 383 & 713 & 817 & 1913 \\
& 12 vs 8 & 675 & 462 & 767 & 1904 \\
& 12 vs 4 & 856 & 367 & 686 & 1909 \\
\multirow{3}{*}{tables}
& 4 vs 8 & 484 & 758 & 670 & 1912 \\
& 12 vs 8 & 709 & 455 & 612 & 1776 \\
& 12 vs 4 & 937 & 379 & 590 & 1906 \\
\bottomrule
\end{tabular}
\caption{Part 1 of Response Counts for Information Richness under Homogeneous Format Conditions }
\label{tab:homo_nums_1}
\end{table}
\begin{table}[htbp]
\centering
\scriptsize
\renewcommand{\arraystretch}{0.9}
\begin{tabular}{lcrrrr}
\toprule
\textbf{Group} & \textbf{Format Pair} & \textbf{Pref-A} & \textbf{Pref-B} & \textbf{Both} & \textbf{Total} \\
\hline
\rowcolor{lightgray}\multicolumn{6}{c}{\textbf{Qwen3-30B-A3B-FP8}}\\
\hline
\multirow{3}{*}{KGs}
& 4 vs 8 & 378 & 561 & 988 & 1927 \\
& 12 vs 8 & 529 & 434 & 963 & 1926 \\
& 12 vs 4 & 580 & 348 & 998 & 1926 \\
\multirow{3}{*}{infoboxes} 
& 4 vs 8 & 401 & 683 & 840 & 1924 \\
& 12 vs 8 & 633 & 484 & 799 & 1916 \\
& 12 vs 4 & 851 & 371 & 697 & 1919 \\
\multirow{3}{*}{tables}
& 4 vs 8 & 525 & 825 & 574 & 1924 \\
& 12 vs 8 & 819 & 537 & 566 & 1922 \\
& 12 vs 4 & 1021 & 415 & 482 & 1918 \\
\hline
\rowcolor{lightgray}\multicolumn{6}{c}{\textbf{Gemma-2-9b-it-FP8}}\\
\hline
\multirow{3}{*}{KGs}
& 4 vs 8 & 445 & 482 & 1017 & 1944 \\
& 12 vs 8 & 468 & 444 & 986 & 1898 \\
& 12 vs 4 & 473 & 433 & 1016 & 1922 \\
\multirow{3}{*}{infoboxes} 
& 4 vs 8 & 434 & 549 & 962 & 1945 \\
& 12 vs 8 & 549 & 518 & 875 & 1942 \\
& 12 vs 4 & 649 & 467 & 822 & 1938 \\
\multirow{3}{*}{tables}
& 4 vs 8 & 552 & 734 & 658 & 1944 \\
& 12 vs 8 & 716 & 549 & 678 & 1943 \\
& 12 vs 4 & 846 & 489 & 603 & 1938 \\
\hline
\rowcolor{lightgray}\multicolumn{6}{c}{\textbf{Gemma-2-27b-it-FP8}}\\
\hline
\multirow{3}{*}{KGs}
 & 4 vs 8 & 256 & 337 & 872 & 1465 \\
 & 12 vs 8 & 394 & 393 & 1190 & 1977 \\
 & 12 vs 4 & 471 & 300 & 1183 & 1954 \\
\multirow{3}{*}{infoboxes} 
 & 4 vs 8 & 323 & 413 & 1257 & 1993 \\
 & 12 vs 8 & 414 & 398 & 1171 & 1983 \\
 & 12 vs 4 & 532 & 351 & 1105 & 1988 \\

\multirow{3}{*}{tables}
 & 4 vs 8 & 436 & 622 & 931 & 1989 \\
 & 12 vs 8 & 622 & 463 & 901 & 1986 \\
 & 12 vs 4 & 742 & 425 & 822 & 1989 \\
\hline
\rowcolor{lightgray}\multicolumn{6}{c}{\textbf{GLM-4-9b-chat-hf}}\\
\hline
\multirow{3}{*}{KGs}
& 4 vs 8 & 472 & 662 & 853 & 1987 \\
& 12 vs 8 & 633 & 538 & 810 & 1981 \\
& 12 vs 4 & 703 & 454 & 828 & 1985 \\
\multirow{3}{*}{infoboxes} 
& 4 vs 8 & 568 & 960 & 451 & 1979 \\
& 12 vs 8 & 898 & 769 & 306 & 1973 \\
& 12 vs 4 & 1085 & 571 & 327 & 1983 \\
\multirow{3}{*}{tables}
& 4 vs 8 & 721 & 1027 & 211 & 1959 \\
& 12 vs 8 & 1005 & 753 & 217 & 1975 \\
& 12 vs 4 & 1164 & 653 & 165 & 1982 \\
\hline
\rowcolor{lightgray}\multicolumn{6}{c}{\textbf{Mistral-7B-Instruct-v0.3}}\\
\hline
\multirow{3}{*}{KGs}
& 4 vs 8 & 360 & 627 & 1007 & 1994 \\
& 12 vs 8 & 530 & 435 & 1027 & 1992 \\
& 12 vs 4 & 675 & 340 & 973 & 1988 \\
\multirow{3}{*}{infoboxes} 
& 4 vs 8 & 368 & 715 & 908 & 1991 \\
& 12 vs 8 & 612 & 441 & 931 & 1984 \\
& 12 vs 4 & 826 & 327 & 831 & 1984 \\
\multirow{3}{*}{tables}
& 4 vs 8 & 573 & 755 & 662 & 1990 \\
& 12 vs 8 & 870 & 537 & 580 & 1987 \\
& 12 vs 4 & 1029 & 422 & 537 & 1988 \\
\bottomrule
\end{tabular}
\caption{Part 2 of Response Counts for Information Richness under Homogeneous Format Conditions }
\label{tab:homo_nums_2}
\end{table}

\begin{table}[htbp]
\centering
\scriptsize
\renewcommand{\arraystretch}{0.9}
\begin{tabular}{lcrrrr}
\toprule
\textbf{Group} & \textbf{Format Pair} & \textbf{Pref-A} & \textbf{Pref-B} & \textbf{Both} & \textbf{Total} \\
\hline
\rowcolor{lightgray}\multicolumn{6}{c}{\textbf{Llama-3.1-8B-Instruct}}\\
\hline
\multirow{3}{*}{KGs}
& 0.45 vs un & 401 & 445 & 1011 & 1857 \\
& 0.9 vs un & 507 & 404 & 942 & 1853 \\
& 0.45 vs 0.9 & 404 & 441 & 1009 & 1854 \\
\multirow{3}{*}{infoboxes} 

& 0.45 vs un & 450 & 534 & 848 & 1832 \\
& 0.9 vs un & 416 & 580 & 844 & 1840 \\
& 0.45 vs 0.9 & 470 & 441 & 935 & 1846 \\
\multirow{3}{*}{tables}
& 0.45 vs un & 521 & 550 & 787 & 1858 \\
& 0.9 vs un & 526 & 551 & 777 & 1854 \\
& 0.45 vs 0.9 & 516 & 502 & 833 & 1851 \\
\hline
\rowcolor{lightgray}\multicolumn{6}{c}{\textbf{GPT-4o-mini}}\\
\hline
\multirow{3}{*}{KGs}
& 0.45 vs un & 374 & 369 & 1140 & 1883 \\
& 0.9 vs un & 397 & 392 & 1093 & 1882 \\
& 0.45 vs 0.9 & 371 & 394 & 1117 & 1882 \\
\multirow{3}{*}{infoboxes} 

& 0.45 vs un & 380 & 439 & 1059 & 1878 \\
& 0.9 vs un & 398 & 488 & 993 & 1879 \\
& 0.45 vs 0.9 & 435 & 431 & 1015 & 1881 \\
\multirow{3}{*}{tables}
& 0.45 vs un & 394 & 408 & 1079 & 1881 \\
& 0.9 vs un & 391 & 398 & 1094 & 1883 \\
& 0.45 vs 0.9 & 400 & 407 & 1040 & 1847 \\
\hline
\rowcolor{lightgray}\multicolumn{6}{c}{\textbf{Qwen3-8B-FP8}}\\
\hline
\multirow{3}{*}{KGs}
& 0.45 vs un & 482 & 504 & 917 & 1903 \\
& 0.9 vs un & 554 & 485 & 751 & 1790 \\
& 0.45 vs 0.9 & 485 & 522 & 925 & 1932 \\
\multirow{3}{*}{infoboxes} 

& 0.45 vs un & 567 & 767 & 589 & 1923 \\
& 0.9 vs un & 79 & 117 & 60 & 256 \\
& 0.45 vs 0.9 & 741 & 667 & 521 & 1929 \\
\multirow{3}{*}{tables}
& 0.45 vs un & 517 & 709 & 703 & 1929 \\
& 0.9 vs un & 572 & 688 & 672 & 1932 \\
& 0.45 vs 0.9 & 604 & 579 & 749 & 1932 \\
\hline
\rowcolor{lightgray}\multicolumn{6}{c}{\textbf{Qwen3-14B-FP8}}\\
\hline
\multirow{3}{*}{KGs}

& 0.45 vs un & 433 & 456 & 1043 & 1932 \\
& 0.9 vs un & 454 & 488 & 991 & 1933 \\
& 0.45 vs 0.9 & 470 & 463 & 999 & 1932 \\
\multirow{3}{*}{infoboxes} 

& 0.45 vs un & 602 & 654 & 666 & 1922 \\
& 0.9 vs un & 606 & 747 & 575 & 1928 \\
& 0.45 vs 0.9 & 620 & 659 & 642 & 1921 \\
\multirow{3}{*}{tables}
& 0.45 vs un & 432 & 560 & 937 & 1929 \\
& 0.9 vs un & 451 & 560 & 922 & 1933 \\
& 0.45 vs 0.9 & 525 & 482 & 926 & 1933 \\
\hline
\rowcolor{lightgray}\multicolumn{6}{c}{\textbf{Qwen3-32B-FP8}}\\
\hline
\multirow{3}{*}{KGs}
 & 0.45 vs un & 453 & 439 & 1022 & 1914 \\
 & 0.9 vs un & 478 & 430 & 1008 & 1916 \\
 & 0.45 vs 0.9 & 437 & 465 & 1013 & 1915 \\
\multirow{3}{*}{infoboxes} 

  & 0.45 vs un & 475 & 534 & 902 & 1911 \\
 & 0.9 vs un & 525 & 566 & 820 & 1911 \\
 & 0.45 vs 0.9 & 518 & 495 & 885 & 1898 \\
\multirow{3}{*}{tables}
 & 0.45 vs un & 507 & 551 & 851 & 1909 \\
 & 0.9 vs un & 465 & 509 & 787 & 1761 \\
 & 0.45 vs 0.9 & 530 & 531 & 853 & 1914 \\
\bottomrule
\end{tabular}
\caption{Part 1 of Response Counts for Structure Quality under Homogeneous Format Conditions}
\label{tab:homo_corrupt_1}
\end{table}

\begin{table}[htbp]
\centering
\scriptsize
\renewcommand{\arraystretch}{0.9}
\begin{tabular}{lcrrrr}
\toprule
\textbf{Group} & \textbf{Format Pair} & \textbf{Pref-A} & \textbf{Pref-B} & \textbf{Both} & \textbf{Total} \\
\hline
\rowcolor{lightgray}\multicolumn{6}{c}{\textbf{Qwen3-30B-A3B-FP8}}\\
\hline
\multirow{3}{*}{KGs}

 & 0.45 vs un & 482 & 472 & 974 & 1928 \\
 & 0.9 vs un & 507 & 484 & 934 & 1925 \\
 & 0.45 vs 0.9 & 446 & 492 & 991 & 1929 \\
\multirow{3}{*}{infoboxes} 

 & 0.45 vs un & 477 & 570 & 874 & 1921 \\
 & 0.9 vs un & 549 & 612 & 759 & 1920 \\
 & 0.45 vs 0.9 & 550 & 563 & 807 & 1920 \\
\multirow{3}{*}{tables}
 & 0.45 vs un & 572 & 616 & 743 & 1931 \\
 & 0.9 vs un & 569 & 634 & 725 & 1928 \\
 & 0.45 vs 0.9 & 564 & 555 & 807 & 1926 \\
 \hline
\rowcolor{lightgray}\multicolumn{6}{c}{\textbf{Gemma-2-9b-it-FP8}}\\
\hline
\multirow{3}{*}{KGs}
 & 0.45 vs un & 463 & 455 & 1020 & 1938 \\
 & 0.9 vs un & 494 & 464 & 987 & 1945 \\
 & 0.45 vs 0.9 & 458 & 474 & 1012 & 1944 \\
\multirow{3}{*}{infoboxes} 

 & 0.45 vs un & 470 & 537 & 941 & 1948 \\
 & 0.9 vs un & 487 & 639 & 817 & 1943 \\
 & 0.45 vs 0.9 & 531 & 529 & 886 & 1946 \\
\multirow{3}{*}{tables}
 & 0.45 vs un & 575 & 583 & 746 & 1904 \\
 & 0.9 vs un & 562 & 642 & 741 & 1945 \\
 & 0.45 vs 0.9 & 594 & 575 & 774 & 1943 \\
\hline
\rowcolor{lightgray}\multicolumn{6}{c}{\textbf{Gemma-2-27b-it-FP8}}\\
\hline
\multirow{3}{*}{KGs}

  & 0.45 vs un & 351 & 409 & 1232 & 1992 \\
 & 0.9 vs un & 374 & 415 & 1198 & 1987 \\
 & 0.45 vs 0.9 & 393 & 381 & 1217 & 1991 \\
\multirow{3}{*}{infoboxes} 
 & 0.45 vs un & 348 & 380 & 1266 & 1994 \\
 & 0.9 vs un & 351 & 395 & 1246 & 1992 \\
 & 0.45 vs 0.9 & 372 & 362 & 1260 & 1994 \\
\multirow{3}{*}{tables}
 & 0.45 vs un & 465 & 538 & 985 & 1988 \\
 & 0.9 vs un & 479 & 559 & 949 & 1987 \\
 & 0.45 vs 0.9 & 514 & 500 & 977 & 1991 \\
\hline
\rowcolor{lightgray}\multicolumn{6}{c}{\textbf{GLM-4-9b-chat-hf}}\\
\hline
\multirow{3}{*}{KGs}
 & 0.45 vs un & 565 & 530 & 889 & 1984 \\
 & 0.9 vs un & 620 & 519 & 846 & 1985 \\
 & 0.45 vs 0.9 & 476 & 533 & 823 & 1832 \\
\multirow{3}{*}{infoboxes} 

 & 0.45 vs un & 558 & 748 & 423 & 1729 \\
 & 0.9 vs un & 666 & 973 & 349 & 1988 \\
 & 0.45 vs 0.9 & 811 & 699 & 469 & 1979 \\

\multirow{3}{*}{tables}
 & 0.45 vs un & 809 & 947 & 228 & 1984 \\
 & 0.9 vs un & 774 & 994 & 211 & 1979 \\
 & 0.45 vs 0.9 & 842 & 820 & 318 & 1980 \\
\hline
\rowcolor{lightgray}\multicolumn{6}{c}{\textbf{Mistral-7B-Instruct-v0.3}}\\
\hline
\multirow{3}{*}{KGs}

 & 0.45 vs un & 532 & 483 & 972 & 1987 \\
 & 0.9 vs un & 635 & 518 & 811 & 1964 \\
 & 0.45 vs 0.9 & 545 & 558 & 873 & 1976 \\
\multirow{3}{*}{infoboxes} 

 & 0.45 vs un & 467 & 554 & 972 & 1993 \\
 & 0.9 vs un & 474 & 679 & 820 & 1973 \\
 & 0.45 vs 0.9 & 585 & 508 & 900 & 1993 \\
\multirow{3}{*}{tables}
 & 0.45 vs un & 559 & 642 & 791 & 1992 \\
 & 0.9 vs un & 583 & 651 & 759 & 1993 \\
 & 0.45 vs 0.9 & 542 & 580 & 869 & 1991 \\
\bottomrule
\end{tabular}
\caption{Part 2 of Response Counts for Structure Quality under Homogeneous Format Conditions}
\label{tab:homo_corrupt_2}
\end{table}

\begin{table}[htbp]
\centering
\scriptsize
\renewcommand{\arraystretch}{0.9}
\begin{tabular}{lcrrrr}
\toprule
\textbf{Group} & \textbf{Format Pair} & \textbf{Pref-A} & \textbf{Pref-B} & \textbf{Both} & \textbf{Total} \\
\hline
\rowcolor{lightgray}\multicolumn{6}{c}{\textbf{Llama-3.1-8B-Instruct}}\\
\hline
\multirow{3}{*}{KGs}
& 4 vs texts & 556 & 1222 & 66 & 1844 \\
& 8 vs texts & 675 & 1085 & 85 & 1845 \\
& 12 vs texts & 672 & 858 & 104 & 1634 \\

\multirow{3}{*}{infoboxes} 
& 4 vs texts & 482 & 1180 & 185 & 1847 \\
& 8 vs texts & 486 & 1160 & 193 & 1839 \\
& 12 vs texts & 546 & 1079 & 220 & 1845 \\

\multirow{3}{*}{tables}
& 4 vs texts & 426 & 1210 & 212 & 1848 \\
& 8 vs texts & 476 & 1117 & 251 & 1844 \\
& 12 vs texts & 532 & 1024 & 282 & 1838 \\
\hline
\rowcolor{lightgray}\multicolumn{6}{c}{\textbf{GPT-4o-mini}}\\
\hline
\multirow{3}{*}{KGs}
 & 4 vs texts & 490 & 1076 & 298 & 1864 \\
 & 8 vs texts & 577 & 894 & 361 & 1832 \\
 & 12 vs texts & 645 & 779 & 437 & 1861 \\
\multirow{3}{*}{infoboxes} 
 & 4 vs texts & 323 & 1325 & 222 & 1870 \\
 & 8 vs texts & 325 & 977 & 260 & 1562 \\
 & 12 vs texts & 324 & 795 & 274 & 1393 \\
\multirow{3}{*}{tables}
 & 4 vs texts & 324 & 1054 & 433 & 1811 \\
 & 8 vs texts & 428 & 948 & 488 & 1864 \\
 & 12 vs texts & 503 & 789 & 572 & 1864 \\
\hline
\rowcolor{lightgray}\multicolumn{6}{c}{\textbf{Qwen3-8B-FP8}}\\
\hline
\multirow{3}{*}{KGs}
 & 4 vs texts & 693 & 1060 & 165 & 1918 \\
 & 8 vs texts & 853 & 857 & 208 & 1918 \\
 & 12 vs texts & 523 & 369 & 120 & 1012 \\

\multirow{3}{*}{infoboxes} 
 & 4 vs texts & 464 & 1352 & 100 & 1916 \\
 & 8 vs texts & 529 & 1250 & 138 & 1917 \\
 & 12 vs texts & 605 & 1169 & 140 & 1914 \\

\multirow{3}{*}{tables}
 & 4 vs texts & 420 & 1341 & 164 & 1925 \\
 & 8 vs texts & 443 & 1075 & 182 & 1700 \\
 & 12 vs texts & 465 & 640 & 181 & 1286 \\
\hline
\rowcolor{lightgray}\multicolumn{6}{c}{\textbf{Qwen3-14B-FP8}}\\
\hline
\multirow{3}{*}{KGs}
 & 4 vs texts & 591 & 1091 & 238 & 1920 \\
 & 8 vs texts & 734 & 902 & 281 & 1917 \\
 & 12 vs texts & 855 & 796 & 265 & 1916 \\

\multirow{3}{*}{infoboxes} 
 & 4 vs texts & 432 & 1288 & 198 & 1918 \\
 & 8 vs texts & 525 & 1163 & 229 & 1917 \\
 & 12 vs texts & 546 & 1114 & 253 & 1913 \\

\multirow{3}{*}{tables}
 & 4 vs texts & 346 & 1279 & 294 & 1919 \\
 & 8 vs texts & 429 & 1177 & 316 & 1922 \\
 & 12 vs texts & 543 & 956 & 421 & 1920 \\
\hline
\rowcolor{lightgray}\multicolumn{6}{c}{\textbf{Qwen3-32B-FP8}}\\
\hline
\multirow{3}{*}{KGs}
 & 4 vs texts & 673 & 978 & 244 & 1895 \\
 & 8 vs texts & 764 & 875 & 270 & 1909 \\
 & 12 vs texts & 807 & 713 & 238 & 1758 \\

\multirow{3}{*}{infoboxes} 
  & 4 vs texts & 384 & 1417 & 107 & 1908 \\
 & 8 vs texts & 475 & 1281 & 145 & 1901 \\
 & 12 vs texts & 516 & 1227 & 163 & 1906 \\
\multirow{3}{*}{tables}
 & 4 vs texts & 369 & 1315 & 220 & 1904 \\
 & 8 vs texts & 484 & 1178 & 244 & 1906 \\
 & 12 vs texts & 600 & 992 & 314 & 1906 \\
\bottomrule
\end{tabular}
\caption{Part 1 of Response Counts for Information Richness under Heterogeneous Format Conditions}
\label{tab:hete_nums_1}
\end{table}

\begin{table}[htbp]
\centering
\scriptsize
\renewcommand{\arraystretch}{0.9}
\begin{tabular}{lcrrrr}
\toprule
\textbf{Group} & \textbf{Format Pair} & \textbf{Pref-A} & \textbf{Pref-B} & \textbf{Both} & \textbf{Total} \\
\hline
\rowcolor{lightgray}\multicolumn{6}{c}{\textbf{Qwen3-30B-A3B-FP8}}\\
\hline
\multirow{3}{*}{KGs}
 & 4 vs texts & 651 & 1108 & 159 & 1918 \\
 & 8 vs texts & 807 & 888 & 219 & 1914 \\
 & 12 vs texts & 841 & 685 & 198 & 1724 \\

\multirow{3}{*}{infoboxes} 
 & 4 vs texts & 392 & 1400 & 101 & 1893 \\
 & 8 vs texts & 452 & 1310 & 145 & 1907 \\
 & 12 vs texts & 515 & 1215 & 180 & 1910 \\

\multirow{3}{*}{tables}
 & 4 vs texts & 434 & 1308 & 176 & 1918 \\
 & 8 vs texts & 526 & 1158 & 229 & 1913 \\
 & 12 vs texts & 684 & 947 & 281 & 1912 \\
\hline
\rowcolor{lightgray}\multicolumn{6}{c}{\textbf{Gemma-2-9b-it-FP8}}\\
\hline
\multirow{3}{*}{KGs}
 & 4 vs texts & 488 & 1335 & 113 & 1936 \\
 & 8 vs texts & 563 & 1227 & 141 & 1931 \\
 & 12 vs texts & 626 & 1163 & 142 & 1931 \\

\multirow{3}{*}{infoboxes} 
 & 4 vs texts & 465 & 1362 & 105 & 1932 \\
 & 8 vs texts & 520 & 1297 & 123 & 1940 \\
 & 12 vs texts & 559 & 1249 & 125 & 1933 \\

\multirow{3}{*}{tables}
 & 4 vs texts & 393 & 1392 & 152 & 1937 \\
 & 8 vs texts & 464 & 1306 & 161 & 1931 \\
 & 12 vs texts & 478 & 1230 & 226 & 1934 \\
\hline
\rowcolor{lightgray}\multicolumn{6}{c}{\textbf{Gemma-2-27b-it-FP8}}\\
\hline
\multirow{3}{*}{KGs}
 & 4 vs texts & 571 & 1130 & 275 & 1976 \\
 & 8 vs texts & 687 & 982 & 304 & 1973 \\
 & 12 vs texts & 794 & 836 & 347 & 1977 \\
\multirow{3}{*}{infoboxes} 
 & 4 vs texts & 470 & 1351 & 162 & 1983 \\
 & 8 vs texts & 542 & 1197 & 242 & 1981 \\
 & 12 vs texts & 653 & 961 & 351 & 1965 \\
\multirow{3}{*}{tables}
 & 4 vs texts & 506 & 1095 & 384 & 1985 \\
 & 8 vs texts & 515 & 1040 & 435 & 1990 \\
 & 12 vs texts & 603 & 822 & 481 & 1906 \\
\hline
\rowcolor{lightgray}\multicolumn{6}{c}{\textbf{GLM-4-9b-chat-hf}}\\
\hline
\multirow{3}{*}{KGs}
 & 4 vs texts & 559 & 1371 & 38 & 1968 \\
 & 8 vs texts & 616 & 1286 & 54 & 1956 \\
 & 12 vs texts & 663 & 1209 & 78 & 1950 \\

\multirow{3}{*}{infoboxes} 
 & 4 vs texts & 444 & 1462 & 63 & 1969 \\
 & 8 vs texts & 523 & 1397 & 48 & 1968 \\
 & 12 vs texts & 600 & 1323 & 47 & 1970 \\

\multirow{3}{*}{tables}
 & 4 vs texts & 433 & 1490 & 46 & 1969 \\
 & 8 vs texts & 547 & 1368 & 57 & 1972 \\
 & 12 vs texts & 674 & 1253 & 46 & 1973 \\
\hline
\rowcolor{lightgray}\multicolumn{6}{c}{\textbf{Mistral-7B-Instruct-v0.3}}\\
\hline
\multirow{3}{*}{KGs}
 & 4 vs texts & 565 & 1322 & 69 & 1956 \\
 & 8 vs texts & 764 & 1132 & 76 & 1972 \\
 & 12 vs texts & 845 & 1015 & 103 & 1963 \\

\multirow{3}{*}{infoboxes} 
 & 4 vs texts & 233 & 1673 & 69 & 1975 \\
 & 8 vs texts & 316 & 1608 & 52 & 1976 \\
 & 12 vs texts & 391 & 1519 & 68 & 1978 \\

\multirow{3}{*}{tables}
 & 4 vs texts & 373 & 1502 & 106 & 1981 \\
 & 8 vs texts & 436 & 1452 & 84 & 1972 \\
 & 12 vs texts & 551 & 1318 & 100 & 1969 \\
\bottomrule
\end{tabular}
\caption{Part 2 of Response Counts for Information Richness under Heterogeneous Format Conditions}
\label{tab:hete_nums_2}
\end{table}

\begin{table}[htbp]
\centering
\scriptsize
\renewcommand{\arraystretch}{0.9}
\begin{tabular}{lcrrrr}
\toprule
\textbf{Group} & \textbf{Format Pair} & \textbf{Pref-A} & \textbf{Pref-B} & \textbf{Both} & \textbf{Total} \\
\hline
\rowcolor{lightgray}\multicolumn{6}{c}{\textbf{Llama-3.1-8B-Instruct}}\\
\hline
\multirow{3}{*}{KGs}
 & texts vs un & 1207 & 556 & 74 & 1837 \\
 & 0.45 vs texts & 535 & 1187 & 119 & 1841 \\
 & 0.9 vs texts & 574 & 1158 & 112 & 1844 \\

\multirow{3}{*}{infoboxes} 
 & texts vs un & 1174 & 494 & 175 & 1843 \\
 & 0.45 vs texts & 446 & 1216 & 185 & 1847 \\
 & 0.9 vs texts & 394 & 1248 & 196 & 1838 \\

\multirow{3}{*}{tables}
 & texts vs un & 910 & 722 & 216 & 1848 \\
 & 0.45 vs texts & 679 & 928 & 237 & 1844 \\
 & 0.9 vs texts & 646 & 905 & 295 & 1846 \\
\hline
\rowcolor{lightgray}\multicolumn{6}{c}{\textbf{GPT-4o-mini}}\\
\hline
\multirow{3}{*}{KGs}
 & texts vs un & 1039 & 459 & 261 & 1759 \\
 & 0.45 vs texts & 397 & 1177 & 281 & 1855 \\
 & 0.9 vs texts & 410 & 1139 & 315 & 1864 \\

\multirow{3}{*}{infoboxes} 
 & texts vs un & 1334 & 328 & 200 & 1862 \\
 & 0.45 vs texts & 297 & 1352 & 220 & 1869 \\
 & 0.9 vs texts & 294 & 1349 & 220 & 1863 \\

\multirow{3}{*}{tables}
 & texts vs un & 847 & 529 & 483 & 1859 \\
 & 0.45 vs texts & 477 & 890 & 498 & 1865 \\
 & 0.9 vs texts & 461 & 899 & 511 & 1871 \\
\hline
\rowcolor{lightgray}\multicolumn{6}{c}{\textbf{Qwen3-8B-FP8}}\\
\hline
\multirow{3}{*}{KGs}
 & texts vs un & 1078 & 682 & 164 & 1924 \\
 & 0.45 vs texts & 627 & 1139 & 159 & 1925 \\
 & 0.9 vs texts & 612 & 1135 & 179 & 1926 \\

\multirow{3}{*}{infoboxes} 
 & texts vs un & 1282 & 423 & 92 & 1797 \\
 & 0.45 vs texts & 320 & 1231 & 84 & 1635 \\
 & 0.9 vs texts & 350 & 1469 & 99 & 1918 \\

\multirow{3}{*}{tables}
 & texts vs un & 1051 & 683 & 196 & 1930 \\
 & 0.45 vs texts & 588 & 1019 & 185 & 1792 \\
 & 0.9 vs texts & 579 & 1140 & 208 & 1927 \\
\hline
\rowcolor{lightgray}\multicolumn{6}{c}{\textbf{Qwen3-14B-FP8}}\\
\hline
\multirow{3}{*}{KGs}
 & texts vs un & 1090 & 601 & 225 & 1916 \\
 & 0.45 vs texts & 582 & 1102 & 232 & 1916 \\
 & 0.9 vs texts & 555 & 1112 & 251 & 1918 \\

\multirow{3}{*}{infoboxes} 
 & texts vs un & 1187 & 424 & 182 & 1793 \\
 & 0.45 vs texts & 413 & 1320 & 181 & 1914 \\
 & 0.9 vs texts & 383 & 1318 & 213 & 1914 \\

\multirow{3}{*}{tables}
 & texts vs un & 961 & 582 & 373 & 1916 \\
 & 0.45 vs texts & 539 & 1014 & 368 & 1921 \\
 & 0.9 vs texts & 489 & 948 & 345 & 1782 \\
\hline
\rowcolor{lightgray}\multicolumn{6}{c}{\textbf{Qwen3-32B-FP8}}\\
\hline
\multirow{3}{*}{KGs}
 & texts vs un & 992 & 683 & 233 & 1908 \\
 & 0.45 vs texts & 619 & 1058 & 225 & 1902 \\
 & 0.9 vs texts & 616 & 1047 & 248 & 1911 \\

\multirow{3}{*}{infoboxes} 
 & texts vs un & 1407 & 381 & 119 & 1907 \\
 & 0.45 vs texts & 372 & 1424 & 113 & 1909 \\
 & 0.9 vs texts & 366 & 1418 & 125 & 1909 \\

\multirow{3}{*}{tables}
 & texts vs un & 904 & 711 & 292 & 1907 \\
 & 0.45 vs texts & 644 & 988 & 275 & 1907 \\
 & 0.9 vs texts & 609 & 1023 & 273 & 1905 \\

\bottomrule
\end{tabular}
\caption{Part 1 of Response Counts for Structure Quality under Heterogeneous Format Conditions}
\label{tab:hete_corrupt_1}
\end{table}

\begin{table}[htbp]
\centering
\scriptsize
\renewcommand{\arraystretch}{0.9}
\begin{tabular}{lcrrrr}
\toprule
\textbf{Group} & \textbf{Format Pair} & \textbf{Pref-A} & \textbf{Pref-B} & \textbf{Both} & \textbf{Total} \\
\hline
\rowcolor{lightgray}\multicolumn{6}{c}{\textbf{Qwen3-30B-A3B-FP8}}\\
\hline
\multirow{3}{*}{KGs}
 & texts vs un & 1103 & 659 & 158 & 1920 \\
 & 0.45 vs texts & 619 & 1106 & 188 & 1913 \\
 & 0.9 vs texts & 664 & 1032 & 216 & 1912 \\

\multirow{3}{*}{infoboxes} 
 & texts vs un & 1449 & 378 & 85 & 1912 \\
 & 0.45 vs texts & 360 & 1473 & 83 & 1916 \\
 & 0.9 vs texts & 331 & 1484 & 101 & 1916 \\

\multirow{3}{*}{tables}
 & texts vs un & 995 & 682 & 237 & 1914 \\
 & 0.45 vs texts & 669 & 1008 & 236 & 1913 \\
 & 0.9 vs texts & 643 & 1010 & 263 & 1916 \\
\hline
\rowcolor{lightgray}\multicolumn{6}{c}{\textbf{Gemma-2-9b-it-FP8}}\\
\hline
\multirow{3}{*}{KGs}
 & texts vs un & 1326 & 487 & 125 & 1938 \\
 & 0.45 vs texts & 448 & 1336 & 153 & 1937 \\
 & 0.9 vs texts & 450 & 1347 & 137 & 1934 \\

\multirow{3}{*}{infoboxes} 
 & texts vs un & 1336 & 477 & 121 & 1934 \\
 & 0.45 vs texts & 466 & 1379 & 90 & 1935 \\
 & 0.9 vs texts & 467 & 1361 & 106 & 1934 \\

\multirow{3}{*}{tables}
 & texts vs un & 1116 & 668 & 156 & 1940 \\
 & 0.45 vs texts & 565 & 1199 & 171 & 1935 \\
 & 0.9 vs texts & 542 & 1248 & 147 & 1937 \\
 \hline

\rowcolor{lightgray}\multicolumn{6}{c}{\textbf{Gemma-2-27b-it-FP8}}\\
\hline
\multirow{3}{*}{KGs}
 & texts vs un & 966 & 666 & 350 & 1982 \\
 & 0.45 vs texts & 576 & 1032 & 376 & 1984 \\
 & 0.9 vs texts & 496 & 1175 & 318 & 1989 \\

\multirow{3}{*}{infoboxes} 
 & texts vs un & 1338 & 477 & 165 & 1980 \\
 & 0.45 vs texts & 418 & 1388 & 173 & 1979 \\
 & 0.9 vs texts & 388 & 1413 & 186 & 1987 \\

\multirow{3}{*}{tables}
 & texts vs un & 841 & 785 & 360 & 1986 \\
 & 0.45 vs texts & 662 & 938 & 381 & 1981 \\
 & 0.9 vs texts & 625 & 959 & 395 & 1979 \\
\hline
\rowcolor{lightgray}\multicolumn{6}{c}{\textbf{GLM-4-9b-chat-hf}}\\
\hline
\multirow{3}{*}{KGs}
 & texts vs un & 1379 & 545 & 43 & 1967 \\
 & 0.45 vs texts & 552 & 1378 & 40 & 1970 \\
 & 0.9 vs texts & 604 & 1317 & 48 & 1969 \\

\multirow{3}{*}{infoboxes} 
 & texts vs un & 1497 & 422 & 47 & 1966 \\
 & 0.45 vs texts & 432 & 1485 & 52 & 1969 \\
 & 0.9 vs texts & 446 & 1473 & 53 & 1972 \\

\multirow{3}{*}{tables}
 & texts vs un & 1229 & 700 & 38 & 1967 \\
 & 0.45 vs texts & 637 & 1296 & 36 & 1969 \\
 & 0.9 vs texts & 646 & 1280 & 40 & 1966 \\
\hline
\rowcolor{lightgray}\multicolumn{6}{c}{\textbf{Mistral-7B-Instruct-v0.3}}\\
\hline
\multirow{3}{*}{KGs}
 & texts vs un & 1343 & 557 & 66 & 1966 \\
 & 0.45 vs texts & 491 & 1419 & 58 & 1968 \\
 & 0.9 vs texts & 477 & 1446 & 53 & 1976 \\

\multirow{3}{*}{infoboxes} 
 & texts vs un & 1692 & 245 & 42 & 1979 \\
 & 0.45 vs texts & 260 & 1667 & 47 & 1974 \\
 & 0.9 vs texts & 269 & 1656 & 54 & 1979 \\

\multirow{3}{*}{tables}
 & texts vs un & 1244 & 654 & 80 & 1978 \\
 & 0.45 vs texts & 645 & 1250 & 84 & 1979 \\
 & 0.9 vs texts & 646 & 1231 & 100 & 1977 \\
\bottomrule
\end{tabular}
\caption{Part 2 of Response Counts for Structure Quality under Heterogeneous Format Conditions}
\label{tab:hete_corrupt_2}
\end{table}

\end{document}